%% file: main.tex
\pdfoutput=1  

\documentclass[]{hdsr}

\graphicspath{{Figure/}}

\input{macros}

\begin{document}

\newgeometry{bottom=1.5in}


\begin{center}

  \title[A Diffusion-Model Subpopulation Digital Twin for Mobile Health Deployment]{A Diffusion-Model Subpopulation Digital Twin for Mobile Health Deployment: A Case Study on the HeartSteps Intervention}
  \maketitle

  \thispagestyle{empty}

  \vspace*{.2in}

  \begin{tabular}{cc}
    Ziping Xu\upstairs{\affilone,*}, Yuyi Chang\upstairs{\affiltwo,*}, Chenshun Ni\upstairs{\affilone}, Nithin Sugavanam\upstairs{\affiltwo},\\
    Asim H. Gazi\upstairs{\affilthree}, Pedja Klasnja\upstairs{\affilfour}, Emre Ertin\upstairs{\affiltwo}, Susan Murphy\upstairs{\affilthree}
   \\[0.25ex]
   {\small \upstairs{\affilone} University of North Carolina at Chapel Hill} \\
   {\small \upstairs{\affiltwo} The Ohio State University} \\
   {\small \upstairs{\affilthree} Harvard University} \\
   {\small \upstairs{\affilfour} University of Michigan}
  \end{tabular}

  \emails{
    \upstairs{*}Ziping Xu and Yuyi Chang are co-first authors. Corresponding author: Ziping Xu (zipingxu@unc.edu).
  }
  \vspace*{0.4in}

\begin{abstract} 
\noindent
Mobile-health interventions increasingly use online learning and decision making algorithms to personalize when to nudge users toward healthier behavior, but a poorly designed algorithm can burden and disengage participants.
New algorithm design decisions should therefore be vetted against realistic simulated users before each real-life deployment.
We propose a method to develop ``JITAI-Twins'': digital twins of a target subpopulation for comparing candidate online algorithms before a just-in-time adaptive intervention (JITAI) deployment.
The method builds on a conditional time-series diffusion model that is \emph{temporally consistent} (future actions do not affect the generated past), and it supports repeated updating from three sources of information, in three steps: pre-training on a large observational dataset, fine-tuning on small prior intervention deployments in related populations, and inference-time calibration to the next target population from domain-scientist expertise.
We validate the twin at each pre-deployment stage of the long-running HeartSteps series (v2 through v4) of physical-activity suggestion intervention deployments, treating each successive deployment as an upcoming study.
The proposed method reproduces the target subpopulation's temporal and between-participant structure better than simpler simulators. These results suggest that our twin can be used to simulate a target deployment before it runs, the prerequisite for testing and informing online algorithm design decisions.

\end{abstract}

\end{center}

\small

\section{Why a Digital Twin for Mobile Health?}
\label{sec:intro}

Mobile health (mHealth) delivers health support through ubiquitous wireless devices such as smartphones and wearables, reaching people in the course of everyday life rather than in the clinic, and at a scale and cost that conventional care cannot match \citep{free2013effectiveness,steinhubl2015emerging,kumar2013mobile}.
mHealth interventions are used to help people achieve desired improvement in  everyday behaviors such as physical activity, medication adherence, and stress management.
Most modern mHealth interventions involve intervention nudges (e.g.,  walking suggestions,  breathing-exercise reminders, or  stress-management suggestions) that are delivered through a mobile app or wearable, and choose for that individual when and whether to send an intervention nudge at each of hundreds or thousands of decision points. 
Interventions that aim to deliver this kind of support  are called just-in-time adaptive interventions (JITAIs) \citep{nahum2016just}.
A growing trend is to personalize JITAIs with an algorithm that learns, from each participant's responses, when and what type of intervention nudge to deliver \citep{gazi2026pJITAIs}.
These are online learning algorithms such as reinforcement learning algorithms \citep{sutton2018reinforcement,lattimore2020bandit} that learn and make decisions on the fly to maximize a reward signal through repeated interaction, prediction algorithms on latent states, or online imputation algorithms imputing missing values adaptively.  
Examples of such deployments span behavioral domains including physical activity (HeartSteps, \citealp{liao2020personalized}), substance use (MiWaves, \citealp{ghosh2024miwaves}), oral self-care (Oralytics, \citealp{trella2025deployed}), medication adherence (the REINFORCE trial, \citealp{lauffenburger2024reinforce}), weight loss~\citep{forman2019reinforcement}, chronic pain~\citep{piette2022aicbtcp}, personalized step goals~\citep{zhou2018stepgoals}, smartphone exercise reminders~\citep{wang2021reminders}, and maternal and child health~\citep{mate2022restless}. 

A major challenge in designing these algorithms is the number of design choices: which features describe the participant's state, how aggressively the algorithm explores, and how much it plans into the future.
A poor combination can waste a participant's attention, increase burden, and compromise the effectiveness of the JITAI.
Practitioners aim to commit to an algorithm design \emph{prior} to each deployment, and to avoid mid-deployment improvisation that may undermine reproducibility \citep{ghosh2025reproducible}.
To make that commitment confidently, one approach is to test many candidate hyperparameter values using a realistic simulator.

Such a simulator is an instance of a \emph{digital twin} \citep{nas2024digitaltwins,niederer2021scaling}.
Following the National Academies' definition adopted by \citet{gazi2025digital}, a digital twin couples a virtual model of a physical system to that system through a two-way exchange: the model is built and continually refreshed from the system's data, and its simulations in turn inform decisions taken in the physical world.
When the simulator's purpose is to test and improve the JITAI and the algorithm that personalizes it, we call the simulator a \emph{JITAI-Twin} \citep{gazi2025digital}.
Unlike the person-specific digital twins familiar from clinical medicine, a JITAI-Twin represents a \emph{subpopulation}: it reproduces how a group of participants behaves, and how that behavior responds to intervention nudges, so that candidate algorithm designs can be compared on it before anyone is enrolled. Here we develop a three-step method for building a JITAI-Twin.
The developed method is illustrated using  a case study on the HeartSteps mHealth intervention, a decade-long series of physical-activity trials.


\begin{figure}[h!]
  \centering
  \includegraphics[width=0.85\textwidth]{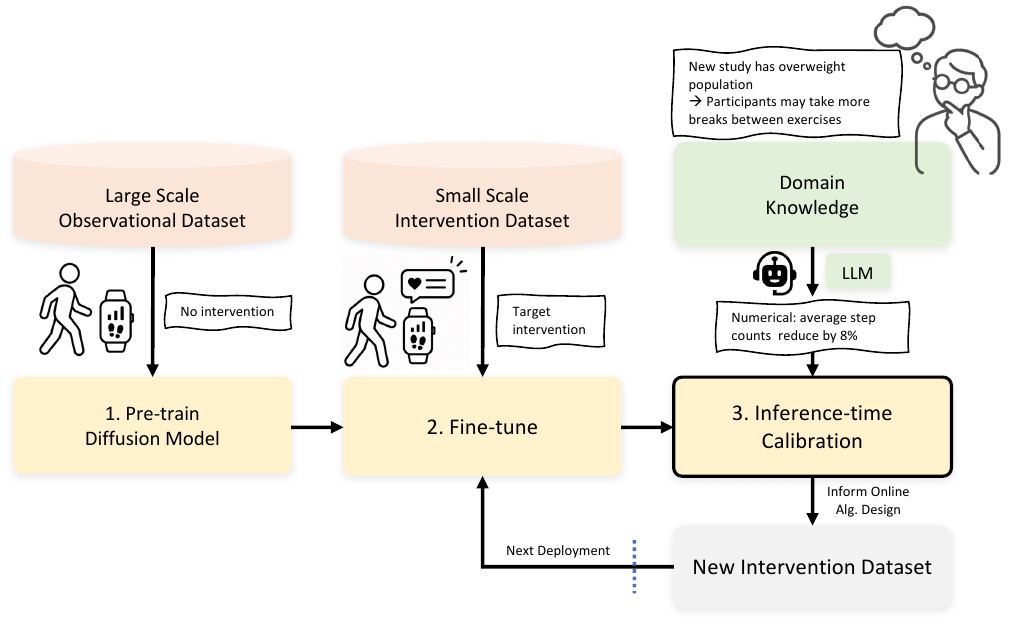}
  \caption{
  The JITAI-Twin pipeline: pre-train on \emph{All of Us}, fine-tune on
  prior HeartSteps deployments, then calibrate the sampler to the target deployment
  at inference time from domain knowledge or an LLM forecast. The fine-tuning data are small-scale \emph{intervention} data from prior deployments, not data from the target population. 
  }
  \label{fig:digital_twin}
  \end{figure}

Building a JITAI-Twin from prior data is challenging for two reasons that recur across mHealth studies \citep{gazi2026statisticalRLreal}.  The first challenge is 
\emph{data scarcity}, due to the targeted subpopulation, which is often costly to recruit. Examples include individuals in cardiac rehabilitation and individuals who recently underwent a blood and marrow transplant.
Thus each deployment of the intervention might involve only 40--80 participants. 
The second challenge is \emph{data impoverishment} \citep{gazi2025digital}. Consecutive deployments of an mHealth intervention might target individuals across different subpopulations, occur in different geographical locations, use different sensing hardware, and unfold in differing societal contexts, so a twin built from the previous deployment's data will need to be updated in preparation for each new deployment. 

\textbf{Overview.}
This paper proposes a method for building and updating JITAI-Twins from  three sources of information: large observational datasets, data from small prior intervention deployments, and domain scientists' expertise.
The proposed approach has three steps (outlined in Figure~\ref{fig:digital_twin}).  The first step  is to \emph{pre-train} a time-series diffusion model \citep{ho2020denoising}, adapted from the TimeWeaver architecture \citep{narasimhan2024time}, on a large observational dataset of people monitored without any intervention, which gives the twin a broad model of common, shared behavioral dynamics (in a physical activity intervention, this might be day-to-day walking behaviors).  The second step is to  \emph{fine-tune} the pre-trained model on prior deployments that included the target intervention nudge with the goal of enabling the twin to simulate how behavior is shifted due to the intervention nudges.
The third step is to \emph{calibrate} the twin for the upcoming deployment using only what is known before that deployment runs: its protocol, and a domain scientist's judgment about how the new population or social context will differ.  Here \emph{calibration} means adjusting the fine-tuned twin so its output better matches the upcoming deployment's population.

We demonstrate the full pipeline using HeartSteps (HS) \citep{klasnja2019efficacy,liao2020personalized}, a long-running series of physical-activity deployments (v2 to v4), by replaying its deployment sequence.
At the first hand-off (before HeartSteps v2/v3 deployment), no intervention data are available for fine-tuning. The calibrated pre-trained twin matches the overall distribution of participant behavior about as well as the simpler baseline simulators.
At the second hand-off (before HeartSteps v4 deployment), the prior deployment supplies intervention data for fine-tuning, and the twin reproduces the behavioral dynamics more faithfully than both the simpler simulators and its uncalibrated counterpart.
At both hand-offs, the calibrated twin better preserves how behavior varies across participants: in its simulations, demographic subgroups differ in the same direction as in the real data, a between-individual structure that the simpler simulators largely ignore.

\textbf{Outline.}
Section~\ref{sec:related} situates the work among prior simulators, time-series diffusion models, and digital twins.
Section~\ref{sec:problem} introduces the HeartSteps deployments and data.
Section~\ref{sec:method} describes the diffusion model and all three steps: pre-training, fine-tuning, and inference-time calibration.
Section~\ref{sec:generation} evaluates the twin on the HeartSteps replay, including how accurate the LLM-predicted calibration target is (Section~\ref{sec:llm}).
Section~\ref{sec:discussion} closes with how the twin is meant to inform the pre-deployment design of online learning and decision making algorithms, why we leave that use to a prospective test, what we did and did not learn, and where the framework should go next.

\subsection{Related work}
\label{sec:related}

\textbf{Simulators for mobile-health algorithm development.}
The dominant approach to testing JITAI algorithms before deployment has been structural simulators fit to a prior deployment: linear state-space or structural-equation models of a participant's next state given context and action, as used in prior HeartSteps testbeds \citep{liao2020personalized}.
These models are cheap and interpretable, but they cannot capture the nonlinear, zero-inflated, and heterogeneous dynamics of real activity data.
A related family builds \emph{mechanistic} models of behavior and uses them to construct simulators. For instance, \cite{mintz2020nonstationary} models explicit habituation and recovery dynamics. \cite{steven2024modeling} adopts a control-system model of engagement and activity. These encode domain structure explicitly, but like the structural simulators above they commit to a parametric form that real activity data may not follow.
A second family resamples prior-deployment data directly, including queue-based replay \citep{mandel2016offline}, per-state rejection sampling \citep{tang2022towards}, and nearest-neighbor replay \citep{wang2022no}, often in the service of offline policy evaluation \citep{yang2022offline}.
Such methods match the marginal activity distribution well by construction, but they lose temporal coherence, splicing real hours together in unrealistic orders.
Our twin departs from both by learning a conditional generative model that reproduces temporal structure and between-participant heterogeneity, not just the marginal.

\textbf{Conditional time-series diffusion.}
Diffusion models were introduced for image generation \citep{ho2020denoising} and adapted to time series for audio synthesis \citep{kongdiffwave}, probabilistic forecasting \citep{rasul2021autoregressive}, and imputation with structured state-space backbones \citep{alcaraz2022diffusion}.
We build on TimeWeaver \citep{narasimhan2024time}, which conditions a state-space diffusion model on heterogeneous metadata, and on related conditional time-series diffusion work \citep{lin2024diffusion}.
Our model-level contribution is to make the time-axis mixing generate strictly forward in time, so the model respects time-ordering.

\textbf{Inference-time calibration.}
Adjusting a simulation model so its output matches a target system from limited data is the classical problem of computer-model calibration and uncertainty quantification \citep{kennedy2001bayesian}. Our calibration step is a generative, inference-time analogue that adapts a fixed model rather than re-estimating its parameters.
Rather than retraining for each target population, we tilt the sampler at inference time, drawing on reward-tilted and guidance-based sampling for diffusion models \citep{moitra2026steering}.
This lets a single fine-tuned twin be calibrated to many target populations through one trust parameter, and it is what makes calibration from pre-deployment knowledge alone practical.

\section{The HeartSteps Program as a Case Study}
\label{sec:problem}

\begin{figure}[h]
    \centering
    \includegraphics[width=0.6\textwidth]{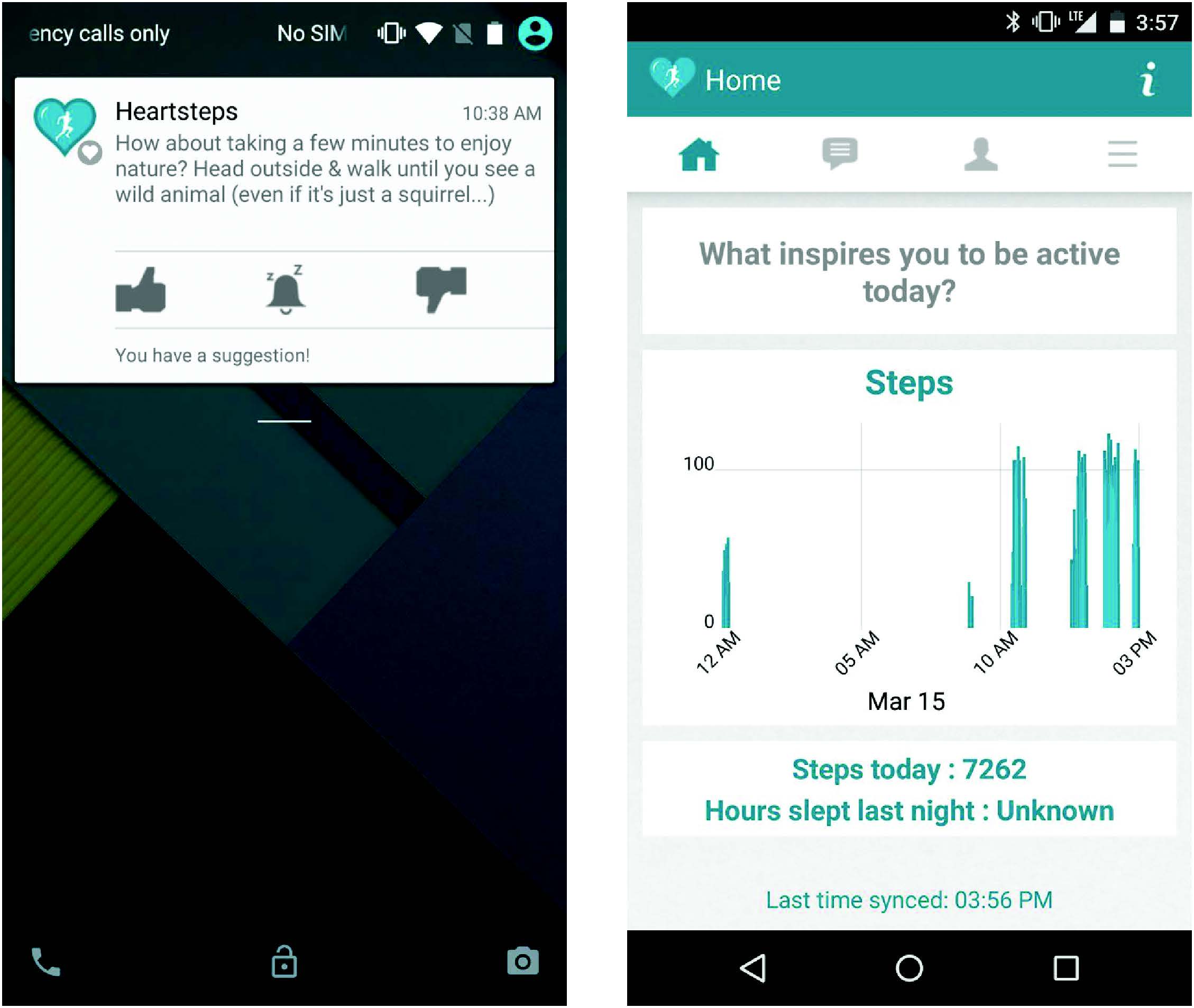}
    \caption{The HeartSteps system: an activity-suggestion notification
    (left) and the step-count graph in the HeartSteps app (right).}
    \label{fig:heartsteps_app}
\end{figure}

HeartSteps is a smartphone app that promotes physical activity through contextually tailored activity suggestions delivered during the day (Figure~\ref{fig:heartsteps_app}).
Behind the app, a reinforcement-learning algorithm decides at each candidate decision point the probability of delivering a contextually tailored activity suggestion, based on the participant's current context (time of day, recent activity, weather, and so on) and the algorithm's evolving belief about which contexts the participant is most receptive to.
The algorithm deployed in HeartSteps is a contextual bandit trained online with Thompson sampling: it maintains a Bayesian model of how the current context predicts a participant's response to the  contextually tailored activity suggestions, and at each decision point randomizes whether to send one in proportion to its current belief that sending will help \citep{liao2020personalized}.
Four micro-randomized trials of HeartSteps, v1 through v4, have run since 2015 \citep{klasnja2019efficacy,liao2020personalized,steven2024modeling}, with study characteristics summarized in Table~\ref{tab:heartsteps_summary}, where we leave out HeartSteps v1 because it used a different intervention nudge. Each successive trial refined the algorithm or the trial protocol.

\begin{figure}[h!]
    \centering
    \includegraphics[width=0.48\textwidth]{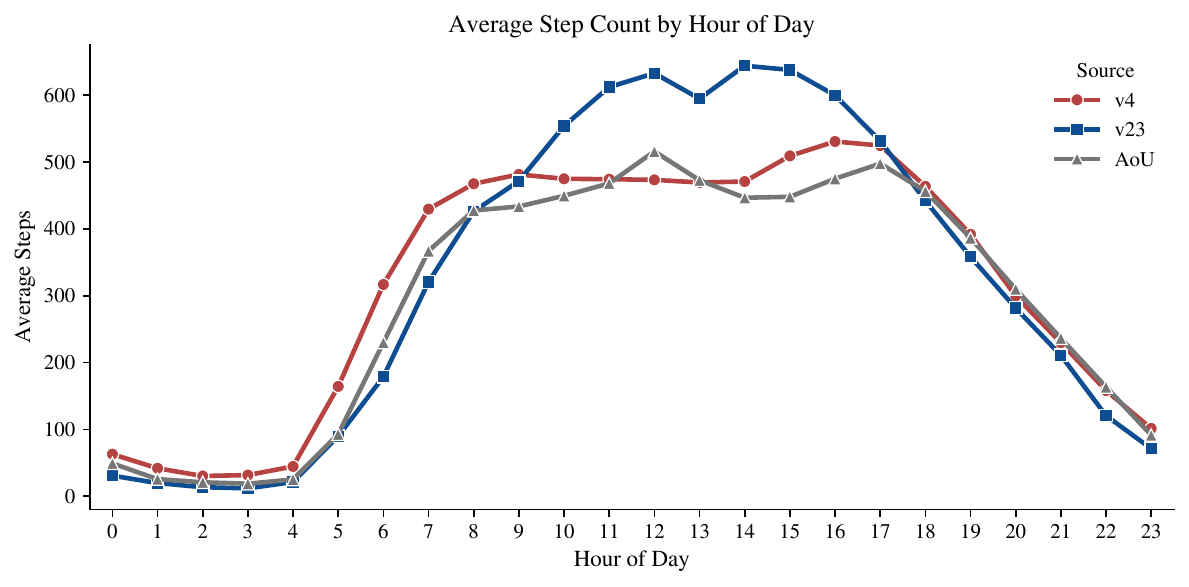}
    \includegraphics[width=0.48\textwidth]{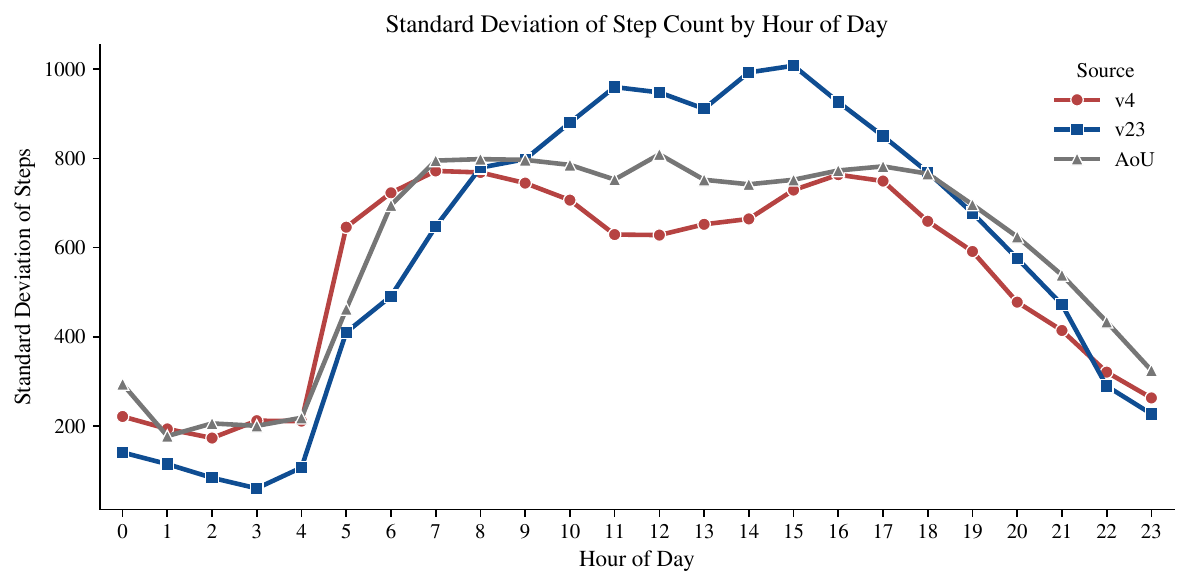}
    \caption{Mean and standard deviation of step-count by hour of day across the HeartSteps deployments and the \emph{All of Us} (AoU) pre-training cohort.
    }
    \label{fig:steps_by_hour}
\end{figure}

The v2 to v4 trials drift through population, geography, and societal context in ways that show up as measurable shifts in how people walk through their day. Trials v2 and v3 followed roughly the same protocol and recruited a similar population (cardiac-rehabilitation patients in Seattle), whereas v4 enrolled a different subpopulation (overweight, sedentary adults) and ran in a different place and time (Southern California, after the onset of the pandemic).
Figure~\ref{fig:steps_by_hour} shows that v2/v3 participants in Seattle, like the national \emph{All of Us} cohort the twin is pre-trained on, exhibit a single mid-day peak around 12--3 PM, while v4 participants in Southern California show a bimodal pattern with peaks in the morning (6--9 AM) and late afternoon (3--6 PM), a plausible signature of post-pandemic hybrid work.

\emph{All of Us} \citep{ramirez2022all} is a national observational cohort run by the NIH: hundreds of thousands of volunteers across the United States contribute health data, and a large subset wears commercial activity trackers in daily life without receiving any intervention.
The roughly $1{,}500$ tracker-wearing \emph{All of Us} participants we pre-train on (Section~\ref{sec:hs_pipeline}) still outnumber any single HeartSteps deployment by more than an order of magnitude, which is what makes \emph{All of Us} a usable substrate for learning baseline activity patterns before the small intervention deployments refine them. 

We replay this decade-long series of HeartSteps deployments to illustrate and evaluate the proposed  JITAI-Twin method. 
As defined above, a digital twin is not a single fixed model but a bi-directional data-model loop: it is continually refreshed as the physical system generates new data.
The HeartSteps series is well-suited to test this loop, because each successive deployment supplies data that updates the twin used to anticipate the next deployment.
We therefore replay the entire sequence of deployments as a case study and treat each successive deployment as if it were an upcoming, not-yet-run study that the twin must anticipate using only the data available up to that point. This yields two pretend-prospective hand-offs, which trace how the twin evolves across the sequence of deployments.
At the first hand-off, we build the twin from only a large external observational dataset, \emph{All of Us}, and treat the v2 and v3 deployments as the upcoming study. At the second hand-off, we refine that twin on the now-available v2/v3 data and treat the most recent deployment, v4, as the upcoming study.

\begin{table}[h]
  \centering
  \caption{Study characteristics across HeartSteps v2--v4.}
  \label{tab:heartsteps_summary}
  \begin{tabular}{l|cccc}
  \hline
  \textbf{Characteristic}  
   & \textbf{v2} & \textbf{v3} & \textbf{v4} \\
  \hline
  Duration (per participant)   
           & 6 months         & 9 months         & 12 months \\
  Location   
     & Seattle, WA      & Seattle, WA      & Southern California \\
  Population 
   & Cardiac patients & Cardiac patients & Overweight \\
  Date range 
       & Jan 2019--       & Oct 2019--       & Jun 2020-- \\
                     & Feb 2020         & Mar 2021         & Oct 2022 \\
  Sample size 
               & 59               & 60               & 79 \\
  \hline
  \end{tabular}
\end{table}

\textbf{Baseline surveys.}
Every participant in these studies completes a baseline structured questionnaire covering demographics, employment, daily routine, psychosocial state, and self-reported physical activity prior to deployment.
This baseline is the only pre-deployment signal the twin has about who a participant is, and it is what lets the twin reproduce how behavior varies from one person to the next.
Previous HeartSteps simulators did not use the data from  baseline questionnaires.
The questionnaires differ across studies (\emph{All of Us}, v2/v3, and v4 each ask overlapping but non-identical questions, in different words and on different answer scales), so there is no fixed feature vector shared across deployments.
We turn this heterogeneous survey into a single conditioning signal by rewriting it as a short natural-language summary and encoding it with a frozen text encoder, which sidesteps the schema-mismatch problem and preserves the semantic content of the questions (Section~\ref{sec:method}, with the full harmonization in Appendix~\ref{appendix:embedding}).

\section{Proposed Digital Twin Approach}
\label{sec:method}

We present a general approach for digital-twin development applicable when there is access to a large observational dataset with longitudinal observations of the targeted behaviors, one or more prior deployments of the target intervention nudges, and domain expertise concerning the population targeted in the upcoming deployment. We use HeartSteps throughout to enhance clarity.

\subsection{The formal setting}
We index deployments by $j = 1, \dots, J$. Deployment $j$ involves $N_j$ participants. Each participant $i$ has baseline survey responses $C_i$ collected before the deployment begins, and produces a discrete-time trajectory of behaviors/states $S_{i,t}$ (in HeartSteps, hourly step counts) and intervention assignments $A_{i,t}$ for $t = 1, \dots, T_j$.
The ``upcoming'' deployment, the one for which we want a JITAI-Twin, is deployment $J$.
A \emph{digital twin} is a model $M$ that, given a virtual participant's baseline $C_i$ and history $\mathcal{H}_t = \{S_{i,\tau}, A_{i,\tau}\}_{\tau \leq t}$, produces a sample of the next state $S_{i,t+1}$.

\subsection{Why a diffusion model} 

The twin must capture three features of the outcome trajectories that simpler simulators routinely miss. First, the marginal distribution of the outcome of interest can take a challenging shape that rules out standard parametric families: in HeartSteps the hourly step count is heavily zero-inflated (people sleep, sit at desks) with a long right tail (a single walk can produce thousands of steps). 
Second, the  outcome trajectory can exhibit strong temporal structure over the modeled window, combining a periodic component with short-range dependence: in HeartSteps weekday mornings look different from Saturday afternoons, and consecutive hours are autocorrelated.
Third, participants' outcome trajectories are heterogeneous (differ from one another) and this heterogeneity is likely only partially explained by baseline characteristics. Further, the baseline characteristics are often encoded in unstructured free-form text.
Conditional diffusion models are well suited to all three: they impose no parametric form on the state transition; they generate a whole window jointly so temporal structure is modeled directly rather than one step at a time, and they admit rich conditioning such as unstructured demographic information and the long history of past treatments.
First proposed for image generation \citep{ho2020denoising} and adapted to time series for forecasting and imputation \citep{rasul2021autoregressive,alcaraz2022diffusion,lin2024diffusion}, they produce samples by gradually denoising Gaussian noise.

\subsection{Architecture at a glance}

We build on \emph{TimeWeaver} \citep{narasimhan2024time}, a state-space conditional diffusion model designed to condition time-series generation on heterogeneous metadata. The generator is a stack of residual blocks, each mixing the target outcome sequence
along time with a state-space kernel and attending to the survey embedding through cross-attention. It produces samples by gradually denoising Gaussian noise, with the survey and the intervention nudge sequence injected at every denoising step. In HeartSteps, our JITAI-Twin generates a participant's hourly step count over a whole week, all $L = 168$ hours at once rather than one hour at a time, conditioned on the baseline survey and the week's intervention sequence. Generating the entire trajectory all at once helps the model capture the complex temporal structure, but does not align with the target formulation described above. We close this gap through ensuring temporal consistency and rolling out the twin through repeated generation while freezing the already generated part. 

\textbf{A temporally consistent generator for evaluating online learning and decision making algorithms.}
Because the twin is used to evaluate online learning and decision making algorithms, its temporal mixing must be strictly forward in time: the $t$-th generation never depends on a future action.
Formally, we require the twin to be \emph{temporally consistent} (non-anticipating): the next state must be conditionally independent of future interventions given the history and the current action,
\[
S_{t+1} \perp\!\!\!\perp A_{t+2}, A_{t+3}, \ldots \mid \mathcal{H}_t, A_t .
\]
This is a property of the architecture, which fixes the direction of information flow.
This is what lets us roll the twin out under a candidate algorithm: at each decision point we freeze the $1$ to $t$ time points already generated, let the algorithm choose the next action ($A_{t+1}$), and generate the remainder of the trajectory ($t+1$ to $T$) conditioned on that action. Concretely, we left-pad \emph{TimeWeaver}'s 1-D convolutions and use forward-only state-space kernels, so no information flows from later hours back into earlier ones \citep{narasimhan2024time}. The full architecture, training objective, and hyperparameters are in Appendix~\ref{sec:architecture_detail}.

\textbf{Encoding the baseline survey.}
The baseline survey enters through a frozen T5 text encoder \citep{raffel2020t5}.
Since the multiple datasets may use different baseline surveys, we match questions across datasets to a common set, rewrite each participant's responses as a short natural-language summary, and encode that sentence into a $512$-dimensional token sequence that conditions the generator through cross-attention.
This harmonization, which lets one model consume surveys with non-identical schemas, is detailed in Appendix~\ref{appendix:embedding}.

\subsection{Generic pipeline}

The pipeline (Figure~\ref{fig:digital_twin}) has three stages, each drawing on a different source of information that a mobile-health program accumulates.  

\textbf{Stage 1: Pre-training on observational data.}
Stage 1 pre-trains the diffusion model on a large observational dataset of people whose target behavior is monitored without the target intervention, which gives the twin a broad prior over day-to-day behavior before any intervention data refine it.
Because these participants are untreated, this stage trains only the parts of the network that generate behavior from the baseline survey: the state-space blocks that carry temporal structure, the cross-attention that reads the survey embedding, and the input and output projections are all trained, while the intervention-injection layers are not included in this round of training, as there is no delivered intervention to condition on. This prior behavior pattern is learned from far more people than any one deployment can enroll, so the twin enters the next stage already knowing the shape of the behavior and how it varies from person to person. See Appendix~\ref{sec:pretrain} for details regarding this stage and its application in the HeartSteps replay.  

\textbf{Stage 2: Fine-tuning on prior randomized deployments.}
Stage 2 fine-tunes the pre-trained model on one or more small prior deployments in which the target intervention was randomized, so the twin learns how a delivered intervention shifts the next-step behavior on top of the observational prior.
This stage inverts what the first one trained: we freeze the state-space backbone that carries the intervention-agnostic behavioral pattern and update 1) the pathways that inject new information on intervention sequences into it, 2) low-rank adapters \citep[LoRA;][]{hu2022lora} on the survey cross-attention, and 3) the input and output projections. Freezing the backbone leaves only a small number of parameters (1.6\%) to update, which is what makes fitting on a small deployment's worth of participants (roughly one hundred in HeartSteps) feasible without overfitting.
The LoRA rank, iteration counts, learning rates, and the exact set of trained modules as used in the HeartSteps replay are in Appendix~\ref{sec:finetune}.

\begin{figure}[htbp]
  \begin{promptbox}[Generic calibration prompt (Step 1: qualitative reasoning)]\small
  ``You are a \emph{[behavioral-science]} researcher who has worked on a long-running series of \emph{[physical-activity nudge]} micro-randomized trials. We are about to launch the \emph{next} deployment of this program in a new city and need to anticipate, before any data are collected, how its participants' \emph{[daily step pattern]} will differ from the cohorts we have already studied. Use only the following:
  \begin{itemize}[leftmargin=1.4em,topsep=2pt,itemsep=1pt]
  \item \emph{[two prior intervention trials of this program, both in Seattle, enrolling adults with coronary artery disease aged $\ge 40$ recruited from a phase~II cardiac-rehabilitation program]};
  \item \emph{[the upcoming deployment in greater Los Angeles, enrolling sedentary adults with body-mass index 25--45, aged 18--65]}.
  \end{itemize}
  Reason from first principles about how the \emph{[location]} and eligibility criteria would shift the \emph{[hourly step-count]} distribution within a typical day, relative to the prior trials. Structure your answer as: (1)~direction of overall \emph{[daily step count]}; (2)~\emph{[diurnal pattern]}; (3)~\emph{[within-hour variability]}; (4)~one-sentence summary. Be specific about hours of day; under $\sim$500 words.''
  \end{promptbox}

  \smallskip
  \begin{promptbox}[Generic calibration prompt (Step 2: numeric target)]\small
  ``You previously predicted how the \emph{next} deployment's step pattern will differ from the prior trials: [\emph{Step 1 response verbatim}]. Here are the observed prior-trial statistics [raw hourly step counts, pooled across all prior-trial participants and days], for reference: [\emph{24-hour table of prior-trial mean, median, std}]. Output the predicted \emph{[diurnal]} shape: \emph{[24]} non-negative numbers giving the fraction of [the day's steps taken in each hour], summing to one. Commit to where activity sits based on your reasoning; do \emph{not} assume the new shape matches the prior cohort. Output format (required, strict): exactly one JSON object on its own line, no commentary, no markdown fence, in this shape: \texttt{\{"shape": [s0, ..., s23]\}}.''
  \end{promptbox}
  \caption{Two-step prompt for producing the target profile $\vmu^\star$ before the HeartSteps v4 deployment runs. Step~1 elicits a qualitative account of how the new population differs from the prior trials; Step~2 turns it into a normalized target profile for the sampler. Bracketed italics mark the study-specific fields, replaced per deployment.}
  \label{fig:generic_calib_prompt}
\end{figure}

\textbf{Stage 3: Inference-time calibration.}
\label{sec:reward_tilt}
After fine-tuning, the twin will primarily  reflect the population it was fine-tuned on rather than the upcoming deployment's population, and often target data are not available to close that gap.
Stage 3 closes this gap through a sampling-time calibration: we calibrate the twin's sampler toward a target behavior profile for the upcoming deployment, changing no model weights and using no data from the target deployment.
Concretely, write $\vs \in \sR^{L}$ for a generated behavior trajectory (in HeartSteps, the $L = 168$ hourly step counts) and $p_\theta$ for the fine-tuned twin's distribution over such trajectories.
Let $\vmu^\star \in \sR^{L}$ denote the target profile, the mean trajectory that the calibrated twin should match (e.g., the per-hour mean of total steps), and let $\vmu^{\mathrm{base}} = \E_{p_\theta}[\vs]$ denote the same statistic under the uncalibrated (base) twin.
Following the framework of \citet{moitra2026steering}, we modify the diffusion sampler to draw from the tilted distribution $\tilde{p}$ in place of $p_\theta$,
\[
\tilde{p}(\vs) \;\propto\; p_\theta(\vs)\,\exp(\gamma\langle \vv, \vs\rangle),
\qquad \vv = \vmu^\star - \vmu^{\mathrm{base}},
\]
where $\gamma \geq 0$ is a tunable ``trust'' parameter.
Implemented as an additive shift inside the reverse diffusion step (Appendix~\ref{appendix:calibration}), the procedure approximately shifts the mean of $\vs$ by $\gamma \vv$ in the model's operating space while leaving the variance roughly unchanged. With $\gamma = 0$ we recover the uncalibrated twin, and larger $\gamma$ shifts more aggressively toward $\vmu^\star$.
We calibrate from inside the generative process rather than shifting the generated samples post hoc; Appendix~\ref{sec:calib_ablation} compares calibration against a post-hoc moment-matching baseline and shows why the in-process approach better preserves the twin's temporal structure.

The target profile starts as a domain scientist's qualitative pre-deployment knowledge of the new population, reasoning from the prior deployments and the upcoming site's eligibility criteria.
In a real deployment this qualitative input is supplied by the study's scientists. For validation in this paper we have a language model play that role, so the whole procedure runs without a human in the loop.
We then turn the qualitative statement into the quantitative profile the sampler consumes by prompting a language model to translate the verbal reasoning into a numeric target profile that the calibration targets. Figure \ref{fig:generic_calib_prompt} shows the prompt used for the HeartSteps case study, where the bracketed parts mark the study-specific fields, replaced per deployment.
This third stage is where our approach departs most from existing practice: pre-training and fine-tuning both adapt the twin to populations we have already measured, whereas calibration adapts it to one we have not yet seen, using only what is knowable before the deployment begins.

Nothing in the sampler ties $\vmu^\star$ to the per-hour mean of the baseline behavior: the tilt in the display above acts on any linear reward $\langle \vv, \vs\rangle$, so the same mechanism admits richer targets, including action-dependent ones.
We defer these generalizations, and their limits, to the discussion (Section~\ref{sec:discussion}).

\section{Evaluation Using a Replay of HeartSteps}
\label{sec:generation}


Having described the JITAI-Twin approach,
we evaluate, using a replay of HeartSteps, how faithfully the twin reproduces a held-out population. We ask this at \emph{both} pretend-prospective hand-offs of the HeartSteps sequence (Section~\ref{sec:problem}). At the first hand-off, the observational pre-trained twin, calibrated based on HeartSteps v2/v3 pre-deployment knowledge, predicts the first intervention deployment (v2/v3). At the second hand-off, the v2/v3-fine-tuned twin calibrated based on HeartSteps v4 pre-deployment knowledge predicts the next deployment (v4). We first provide some details about how the generic pipeline is specialized to HeartSteps, then describe what we generate and how we evaluate it. 

\subsection{Pipeline details for HeartSteps}
\label{sec:hs_pipeline}

We now describe each stage of the pipeline (Section~\ref{sec:method}) for the replay of the HeartSteps case study (Section~\ref{sec:problem}).

\textbf{Stage 1 (pre-training).}
We use the NIH \emph{All of Us} Research Program as the observational dataset.
\emph{All of Us} shares Fitbit data for roughly $59{,}000$ participants.
We restrict to participants with at least four weeks of Fitbit data and complete responses to the $51$ survey questions used by the participant embedding (Appendix~\ref{appendix:embedding}), and we pre-train on the pre-pandemic window of their data, which yields a pre-training cohort of $1{,}507$ participants (Appendix~\ref{sec:pretrain}).
Days whose total step count does not exceed $200$ are treated as non-wear and masked from the training loss, and hourly values above $45{,}000$ steps are removed as device artifacts.
\emph{All of Us} participants do not receive the HeartSteps contextually tailored activity suggestions; conceptually, pre-training on this cohort provides a ``prior distribution'' over day-to-day step-count trajectories rather than a record of intervention response.

\textbf{Stage 2 (fine-tuning).}
We fine-tune the \emph{All of Us}-pretrained twin on HeartSteps v2 and v3, and the v2/v3-fine-tuned twin on v4, at the two pretend-prospective hand-offs. 

\textbf{Stage 3 (calibration).}
For HeartSteps the calibration target $\vmu^\star$ is the per-hour mean step-count profile of the upcoming deployment, forecast before the deployment by the two-step scientist-and-LLM protocol (Figure~\ref{fig:generic_calib_prompt}, evaluated in Section~\ref{sec:llm}).
We apply the reward-tilted sampler of Section~\ref{sec:reward_tilt} with a single trust parameter per hand-off ($\gamma = 1.0$), treated as fixed a default value.
We calibrate once at each of the two pretend-prospective hand-offs (Section~\ref{sec:problem}): at the first we calibrate the \emph{All of Us}-pretrained twin toward the upcoming v2/v3 deployment, and at the second we calibrate the v2/v3-fine-tuned twin toward v4.

\subsection{What we generate}
For each participant we take their baseline survey and the intervention sequence they actually received during the deployment, and ask the twin to generate four weeks of hourly step counts under that same sequence.
Since the action channel is introduced only at fine-tuning, the AoU-pretrained twin generates from the survey alone, and the recorded action sequence plays no role. The baseline simulators we compare against (Section~\ref{sec:fidelity}) never use the survey: they are state-action transition models fit, at each hand-off and rolled forward under each participant's recorded action sequence.

At each hand-off we evaluate the twin and every baseline against the \emph{entire} target deployment.
Because each candidate's generated weeks form a distribution over synthetic participants, we compare that distribution to the full real cohort at the cohort level. To measure the uncertainty, we randomly sample 30 participant in each batch and compute the metrics. The standard error is computed against 200 such resamples.

\subsection{How we evaluate}
\label{sec:eval_metrics}

We score the twin on five metrics, grouped by the aspect of fidelity each one probes: one for overall distributional fit, two for within-participant temporal structure, and two for between-participant heterogeneity. Full details are deferred to Appendix~\ref{appendix:metrics_baselines}. 

\noindent\textbf{Overall distributional fit.}
\begin{itemize}
  \item The \textbf{daily-profile Wasserstein distance} treats each participant-day as a single point, the 24-vector of that day's hourly step counts on the log scale, and computes the exact Wasserstein distance between the synthetic and real distributions of these daily profiles (all-zero non-wear days excluded).
It measures whether the twin produces whole days that are distributed like real days, in magnitude and in within-day placement at once.
\end{itemize}
\textbf{Within-participant temporal structure (two metrics).}
\begin{itemize}
  \item  These measure whether each participant's hours hang together realistically over time. The \textbf{short-lag autocorrelation MAE} (lags 1--6, averaged over participants) measures how closely the synthetic hour-to-hour correlation matches the real one. The \textbf{active-period run-length Wasserstein} (thresholding each hour at 100 steps) measures whether the twin produces realistic stretches of activity.
\end{itemize}
\textbf{Between-participant heterogeneity (two metrics).}
\begin{itemize}
  \item A heterogeneity ratio measures whether the synthetic participants differ from one another in a v4-like way: we take a per-participant statistic, compute its standard deviation across the synthetic participants, and divide by the same quantity in the real dataset, so that a ratio of one is ideal and a ratio below one means the synthetic participants are too alike. We report this ratio for the two within-participant statistics defined above: \textbf{short-lag autocorrelation} (\textbf{bu-acf}) and \textbf{mean active run length} (\textbf{bu-run}).
\end{itemize}

Formal definitions of all metrics and a description of every baseline simulator used as a comparator are provided in Appendix~\ref{appendix:metrics_baselines}.

\subsection{Fidelity of the twin and the full pipeline}
\label{sec:fidelity}

We replay the two hand-offs in deployment order: the v2/v3 hand-off (panel (a) of Table~\ref{tab:eval_main}) first, then the v4 hand-off (panel (b)).
In each, the twin is compared against three state-action baselines, \emph{KNN}, \emph{per-state rejection sampling (PSRS)} \citep{tang2022towards}, and a \emph{linear SEM} (the structural model used in prior HeartSteps testbeds), none of which condition on the survey. Each rolls forward closed-loop from the previous step, the clock, and the action, with a bootstrap standard error over participants (Appendix~\ref{appendix:metrics_baselines}).

\input{Figure/eval/main_table}

\textbf{Reading the reference rows.}
Each panel closes with two reference rows: the real source data the twin was trained on ($^{\ddagger}$; \emph{All of Us} in panel (a), v2/v3 in panel (b)) and the target deployment's own real data ($^{\S}$).
Both are scored with the same metrics and the same sample-size conventions as the synthetic candidates so the numbers are directly comparable (Appendix~\ref{appendix:metrics_baselines}).
The last row therefore scores finite samples of the target against the target itself, so its distances stay above $0$ and its ratios fluctuate around $1$.

\textbf{The first hand-off (panel (a) of Table~\ref{tab:eval_main}).}
At the first hand-off the operative pre-deployment twin is the AoU-pretrained model calibrated toward v2/v3, with no HeartSteps v2/v3 fine-tuning data (v2/v3 is itself the upcoming deployment).
Its strength is between-participant heterogeneity: from observational pre-training alone it reproduces the v2/v3 autocorrelation spread better than any baseline (bu-acf $0.41$ calibrated, $0.44$ uncalibrated, against at most $0.17$ for the baselines).
On the daily-profile distribution, by contrast, the pre-deployment twin holds no advantage: because the \emph{All of Us} source already sits on top of v2/v3, a resampler of the source produces realistic daily profiles (PSRS reaches a daily Wasserstein of $9.65$, against the floor of $7.93$, the Wasserstein computed on the real v2/v3 data).
Calibration narrows this gap, improving the daily-profile fit ($10.82 \to 10.22$) as it shifts the twin toward the target's activity level, alongside gains in the temporal bout structure (run-length Wasserstein $2.34 \to 1.94$) and run-length heterogeneity (bu-run $0.20 \to 0.25$). 
What calibration cannot supply here is the within-day structure itself: with no fine-tuning on a real deployment, the run-length and daily-profile columns stay comparable to the resampling baselines, and the v2/v3 \emph{oracle} (fine-tuned on v2/v3) fixes this by improving all of the run-length Wasserstein, bu-run, and the daily Wasserstein.

\textbf{The second hand-off (panel (b) of Table~\ref{tab:eval_main}).}
At the v4 hand-off the full pipeline is available, and the calibrated v2/v3 twin (the ``+ calibration'' row) is our operative pre-deployment model.
It is the only non-oracle simulator faithful on all three axes at once: it brings the daily-profile Wasserstein to $8.34$, at the HS v4 real data floor of $8.30$ and below every baseline; it posts the panel's best temporal bout structure; and it preserves the between-participant heterogeneity that every baseline collapses (bu-acf and bu-run at $0.28$ and $0.31$).
The retrieval and parametric baselines instead miss either the daily profile (PSRS and the linear SEM sit at $10.81$ and $10.17$) or the autocorrelation (KNN at ACF-MAE $0.22$), and all of them erase the participant-to-participant spread.
Because these baseline methods were fit on v2/v3, the resamplers do not transfer to v4: PSRS falls from the panel-(a) runner-up on the daily profile to the panel's worst here ($10.81$ against the twin's $8.34$), separated well beyond the error bars.
KNN alone stays within about one standard error of the twin on this single column ($8.62$), while missing the autocorrelation and collapsing heterogeneity, so no baseline is faithful on more than one axis.
Crucially, this is achieved with no v4 data: the calibrated pipeline matches the oracle that does see v4 (within one standard error on every column), so a fully pre-deployment twin is about as faithful as the best-case retrospective one.



\begin{figure}[h!]
\centering
\includegraphics[width=0.98\textwidth]{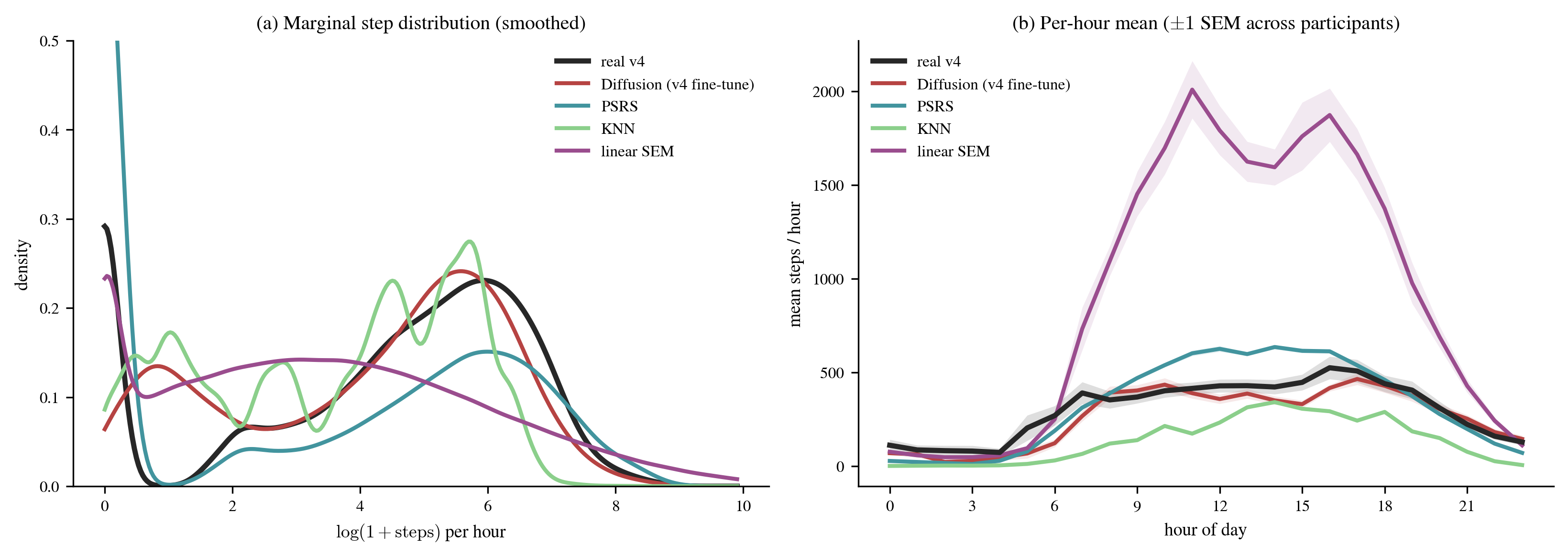}
\caption{Distributional fit on HS v4: the v4-fine-tuned diffusion twin against the Table~\ref{tab:eval_main} baselines and real v4. (a) Marginal $\log(1+\text{steps})$ distribution, as a smoothed kernel density. (b) Per-hour mean, with a $\pm 1$ standard-error-of-the-mean band across participants. 
}
\label{fig:eval_dist}
\end{figure}

Figure~\ref{fig:eval_dist} shows the distributional picture against the baselines for the second hand-off, using the v4-fine-tuned twin (the oracle row of Table~\ref{tab:eval_main}b): it tracks v4's active-hour distribution and long upper tail (panel a) and follows the bimodal diurnal shape in the per-hour mean (panel b), its main miss being the exact-zero spike, which a continuous sampler cannot emit.
The baselines fail differently: PSRS over-produces zeros and overshoots the mid-day mean, KNN collapses toward an over-smooth low-activity profile, and the parametric linear-SEM baseline over-disperses, overshooting the daytime mean severalfold.

\subsection{Capturing demographic conditioning}
\label{sec:demo}

\begin{figure}[h!]
\centering
\includegraphics[width=0.95\textwidth]{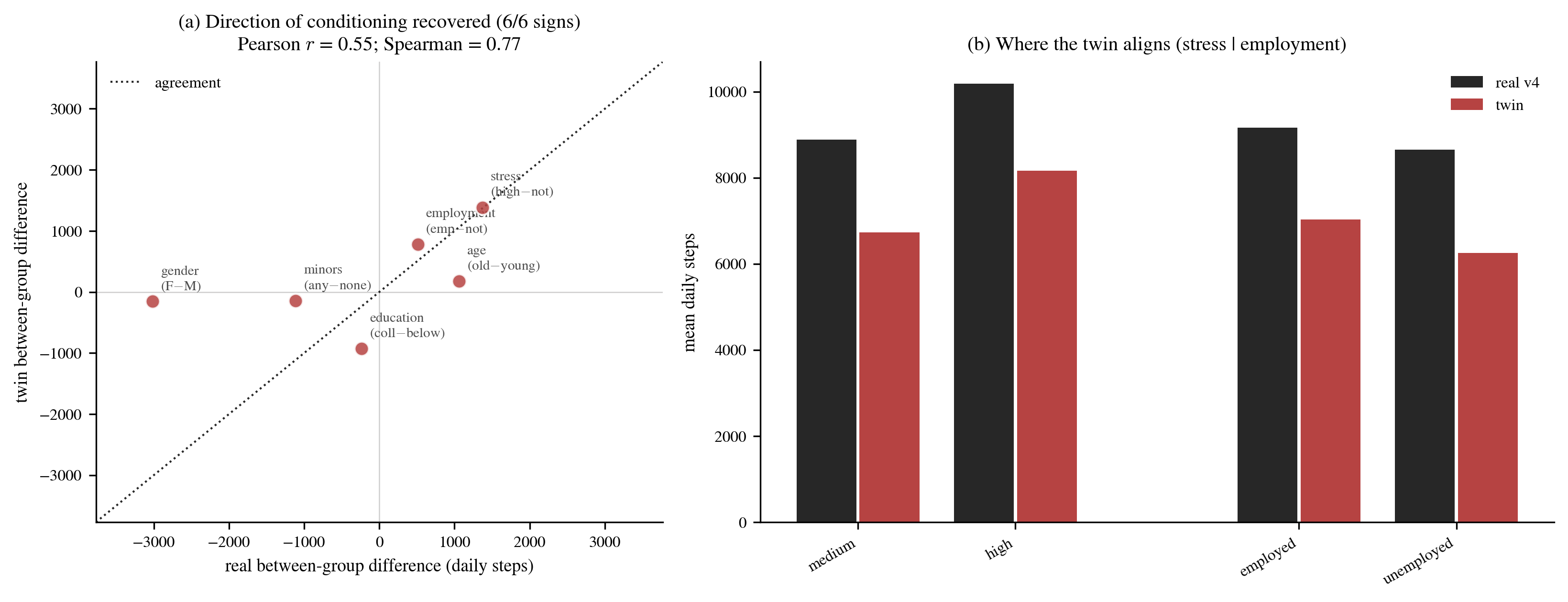}
\caption{Demographic conditioning on v4.
(a) Each point is one demographic variable, binarized into two groups over the
whole cohort; axes are the between-group difference in mean daily steps
(real $x$, twin $y$). All six variables agree in sign.
(b) Two fields the twin matches in direction and magnitude: stress and
employment.}
\label{fig:demographic}
\end{figure}

A useful twin must reproduce not only the pooled population but how behavior \emph{varies} across populations.
We test this by binarizing each survey field into two groups over the whole v4 cohort (six variables: gender, woman vs.\ man; employment, employed vs.\ not; education, college vs.\ below; self-reported stress, high vs.\ not; minors in the household, any vs.\ none; and age, split at the median) and comparing the between-group difference in mean daily steps for the real cohort and for the calibrated twin's generated participants.
The twin recovers the \emph{direction} of every demographic contrast (Figure~\ref{fig:demographic}a): the sign of the between-group difference matches real v4 for all six variables.
On some fields it also recovers the magnitude (Figure~\ref{fig:demographic}b): higher-stress participants walk more, by about the right amount (roughly $+1{,}390$ twin vs.\ $+1{,}370$ real daily steps), and employed participants walk more than unemployed ones.
Across the six variables the between-group differences correlate at Pearson $r = 0.55$ (Spearman $0.77$).
Appendix~\ref{appendix:eda} (Figure~\ref{fig:eda_conditioning_gen}) overlays the generated and real subgroup-mean hourly curves for every field across all three datasets, showing that subgroup ordering and peak timing, not only the sign of the daily gap, carry over.

\subsection{What the twin learned about the intervention}
\label{sec:action_cond}

Demographic conditioning is one half of what the twin must reproduce. The other is how the outcome trajectory responds to the \emph{intervention} nudges. We evaluate what the v4-fine-tuned twin learned by generating on all of the HS v4 participants each week twice, once with a realistic week of suggestions (2.7 times per week in the real HS v4 dataset) and once with no nudges, and evaluating the difference.
The  effect of the nudges is small. 
At the v4 observed send rate, delivering the suggestions changes a participant's weekly step total by only about $-0.02\%$ on average, compared with the steps generated under the all-zero intervention sequence. 

This near-null effect is a property of the HeartSteps program, not of the twin: where a scientist expects a stronger effect than the prior deployments revealed, that expectation can be injected as an action-dependent calibration target, a generalization we defer to the discussion (Section~\ref{sec:discussion}).


\begin{figure}[htbp]
  \centering
  \includegraphics[width=\textwidth]{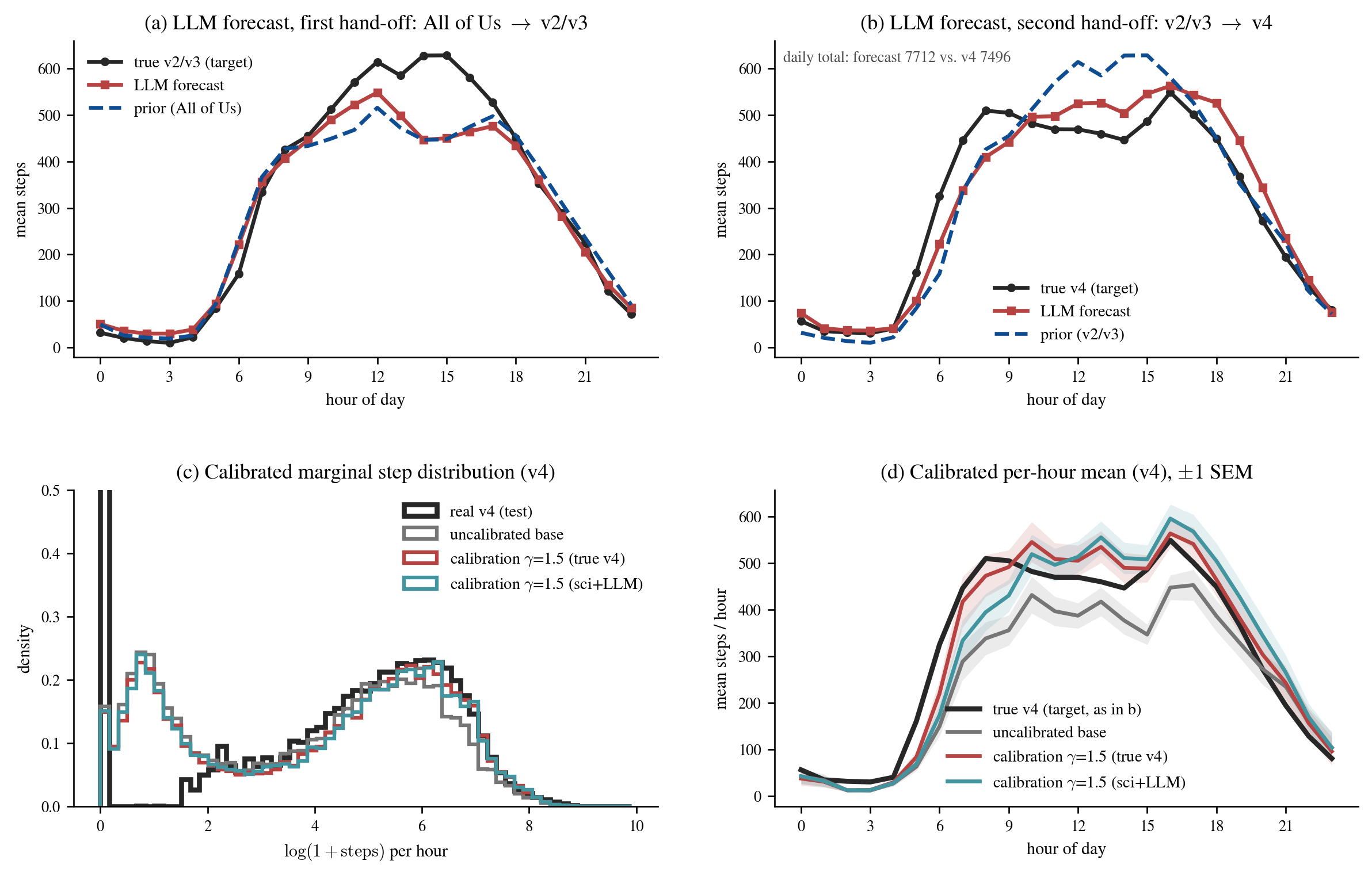}
  \caption{Producing and applying the calibration target.
  \textbf{(a,b)}~The protocol-only LLM forecast (red) of the target diurnal shape at the two hand-offs, against the true target (black) and the prior deployment (AoU for the first hand-off, v2/v3 for the second hand-off); the daily level is taken from the prior.
  \textbf{(c,d)}~Calibrating the v2/v3 twin toward the v4 target moves it onto real v4: it matches the marginal step distribution (c) and, in the per-hour mean (d), tracks the true v4 target (black, the same target as in panel (b)) far more closely than the uncalibrated base, whether calibrated toward the true v4 mean (red) or the pre-deployment scientist-and-LLM forecast (teal). Bands in (d) are $\pm 1$ standard error of the mean across participants; hour of day runs 0--23 from midnight. 
  }
  \label{fig:llm_target}
  \label{fig:calib_steering}
  \end{figure}

\subsection{Evaluating the LLM-predicted calibration target}
\label{sec:llm}

The calibrated pipeline evaluated above relies on a target profile $\vmu^\star$ that must be forecast before the deployment. We now ask how accurate that forecast is.
For HeartSteps the target is the per-hour mean step-count profile, produced by the two-step LLM protocol of Section~\ref{sec:reward_tilt} (Figure~\ref{fig:generic_calib_prompt}): a domain scientist (or an LLM standing in) predicts the qualitative shift of the new population, and an LLM turns that reasoning into a normalized diurnal shape, which we scale by the prior deployment's daily level.
We evaluate the forecast at both pretend-prospective hand-offs.


To keep the replay honest and reproducible, the prompt never names the target deployment or the program and includes no post-deployment results.
The prompt supplies only pre-deployment knowables (the \emph{All of Us} cohort, the two prior Seattle trials with their entry criteria, and the upcoming greater-Los-Angeles site with its own eligibility criteria), from which the model reasons about how location and eligibility shift the hourly step distribution (Step~1, Figure~\ref{fig:generic_calib_prompt}) before committing to a normalized diurnal shape (Step~2, Figure~\ref{fig:generic_calib_prompt}).
Because generation is not deterministic, we draw $N=8$ independent forecasts and take the per-hour median of their normalized shapes, which is stable across draws even as the prose varies.
We run this two-step workflow end to end with GPT-5.4-mini (Azure OpenAI) on both hand-offs (full prompts in Appendix~\ref{appendix:llm}), with the language model standing in for the domain scientist throughout, so the results below are a conservative, fully-automated test rather than a test of what a scientist in the loop could achieve.

Using the v2/v3 to v4 hand-off as a showcase, we first ask whether the model's \emph{qualitative} reasoning (Step~1) recovers the right information about the target population.
It correctly anticipates that the Los Angeles cohort is less active than the Seattle cardiac-rehab participants and that its activity is weighted later in the day.
Its main error is the morning: it predicts a \emph{muted} morning ramp, whereas v4 in fact has a pronounced early-morning peak, so the reasoning is right about the afternoon weighting but wrong about the morning.

Translated into a numeric shape and scaled by the prior's daily level, the \emph{quantitative} forecast then matches v4 on \emph{both} dimensions (Figure~\ref{fig:llm_target}b).
Because the shape was elicited in normalized form, free of the prior deployment's hourly reference, the predicted profile pulls off the prior's afternoon-heavy peak toward v4's later peak; consistent with the qualitative error above, it under-represents v4's early-morning peak.
At the earlier hand-off (Figure~\ref{fig:llm_target}a) the forecast tracks true v2/v3 through the morning ramp but understates its sustained afternoon plateau, staying close to the naive \emph{All of Us} prior. Because that prior already sits near v2/v3, there is little shape gap to close.

Calibrating the twin toward this pre-deployment forecast closes the level gap to v4 and recovers its bimodal diurnal shape (Figure~\ref{fig:calib_steering}c,d), in both the marginal step distribution and the per-hour mean, and does so almost as well as calibrating toward the true v4 mean.
A fuller comparison, including a post-hoc moment-matching baseline and results at both hand-offs, is in Appendix~\ref{sec:calib_ablation}.

\subsection{What the evaluation shows}
\label{sec:takeaways}

Read together, the two hand-offs and the demographic and intervention checks point to three take-home messages about when, and why, a twin like ours earns its cost.

\textbf{The value of the twin tracks the shift between deployments.}
When the source population already sits close to the target, a simple resampler of the source produces realistic daily profiles nearly for free: at the first hand-off PSRS reaches a daily Wasserstein of $9.65$, ahead of the pre-deployment twin's $10.22$, and only the fine-tuned oracle overtakes it.
When the deployments differ, the resamplers no longer track the target and the fine-tuned, calibrated twin does: at the second hand-off the twin sits at the real dataset floor ($8.34$ against the floor's $8.30$) while the same resampler falls to the panel's worst ($10.81$).
So a resampler is a strong baseline when consecutive deployments are similar, whereas the twin's fine-tuning and calibration are what transfer when they differ.

\textbf{When a simpler simulator suffices.}
The same comparison says when the twin is not worth its cost.
If the upcoming deployment resembles the last one and the design decision turns only on the distribution of activity, a resampler such as PSRS is a sensible choice: it is training-free, interpretable, and within reach of the fine-tuned oracle on distributional fit at the first hand-off.
The twin earns its cost when either condition fails: when the populations shift, as at the v4 hand-off, or when the decision consumes structure that a single shared resampling kernel cannot represent, above all the between-participant heterogeneity (bu-acf, bu-run) that every baseline collapses and that choices such as how much to pool across participants turn on.

\textbf{Fine-tuning and calibration divide the labor.}
The two adaptation stages are complementary rather than substitutable (Appendix~\ref{appendix:credit}): fine-tuning supplies the temporal bout structure and well-scaled heterogeneity that calibration alone cannot manufacture, and calibration sets the marginal level on top of a fine-tuned base without disturbing that structure.

\section{Discussion}
\label{sec:discussion}

\subsection*{What this paper adds} The paper makes three
contributions.
(1) A temporally consistent conditional time-series diffusion architecture for digital twins of mobile-health deployments, modified
to make the time-axis state-space kernels forward-only and to inject the intervention sequence through a dedicated forward-only module in each residual block.
The forward-only, non-anticipating path is what lets the twin serve as a valid environment for online learning and decision making algorithms, generating baselines that never leak information from future actions.
(2) An inference-time calibration step that calibrates the twin toward an upcoming deployment using only pre-deployment knowledge.
(3) An evaluation of the first two contributions using an end-to-end replay of the HeartSteps program.
Specifically, the fine-tuned twin is the only simulator that reproduces a held-out HeartSteps deployment's temporal structure and between-participant heterogeneity together with a faithful marginal; each of the simpler baselines sacrifices at least one of the three.   The calibration closes most of the marginal gap and, unlike a per-hour rescaling, preserves the twin's activity-bout structure. The scientist's forecast, drafted by an LLM that predicts the diurnal \emph{shape} while the daily \emph{level} is taken from prior-deployment data, reaches both the level and the shape of an oracle target (per-hour-of-day mean MAE $53$ vs.\ $48$) using no target-deployment data.
The evaluation on HeartSteps shows that the twin's advantage over resampling and structural baselines is exactly the between-participant heterogeneity and temporal structure that decisions such as feature selection or regularization in online algorithms depend on,
and that this advantage is largest when the upcoming deployment differs from the ones already seen.

\subsection*{Using the twin to inform online-algorithm design}
A high-fidelity JITAI-Twin of a subpopulation is critical for informing the online-algorithm design decisions prior to deployment: improvising mid-deployment compromises reproducibility, and a poor configuration can reduce therapeutic effects and burden participants \citep{ghosh2025reproducible}.
Two deployed algorithms illustrate the kinds of decisions the twin is meant to support.
HeartSteps delivers walking suggestions with a contextual bandit trained online by Thompson sampling \citep{thompson1933likelihood,russo2018tutorial,liao2020personalized}; its prior means and variances and its assumed observation noise set how strongly early decisions lean on pooled population beliefs, and ranking such settings requires a simulator that preserves how much participants genuinely differ, exactly the between-participant heterogeneity that every baseline collapses (Section~\ref{sec:takeaways}).
Oralytics, a JITAI promoting oral self-care, decides \emph{when} to place each prompt: the deployed algorithm learns an online, participant-specific model of the next brushing time and schedules the prompt ahead of the predicted time by an offset that grows with the prediction's uncertainty \citep{gazi2025sigmascheduling,gullapalli2026learning}.
Its pre-deployment choices, e.g., the class of behavior-time predictor and the population priors, succeed only when prompts land shortly before the actual brushing events, so rehearsing these choices requires a simulator that reproduces when, and how consistently, each participant acts from day to day.



\subsection*{Generalizing the calibration target}
Here we restricted the calibration target $\vmu^\star$ to the per-hour mean: the tilt of Section~\ref{sec:reward_tilt} acts on any linear reward $\langle \vv, \vs\rangle$, so the same mechanism admits richer targets whenever the quantity a scientist wants to match is approximately linear in the generated trajectory $\vs$.
Two generalizations are worth naming.
First, the target can be made \emph{action-dependent}.
Writing it as $\vmu^\star(a)$ for the intervention sequence $a$ the twin is rolled out under, and tilting by $\vv(a) = \vmu^\star(a) - \vmu^{\mathrm{base}}(a)$, lets a scientist encode a belief about the intervention's effect directly into the sampler: for example, a scientist who expects a positive proximal effect sets $\vmu^\star(a)$ above its no-nudge value $\vmu^\star(\mathbf{0})$ at treated hours, so the calibrated twin lifts the generated steps following a nudge by the elicited amount.
Second, the target need not be a mean.
Any statistic that is approximately linear in $\vs$, such as a higher week-to-week variance or a steeper weekend dip, can be written as its own $\vv$ and tilted toward.
Both are linear-reward tilts the current sampler supports in principle, and we leave their specification and validation to future work.

\subsection*{Limitations}
First, we exercise only one direction of the digital twin's defining two-way coupling.
We demonstrate the data-to-model direction in full, refreshing and re-calibrating the twin as each deployment's data arrive, but the model-to-decision direction is exercised only in simulation replay and not in a real-life deployment (Section~\ref{sec:intro}). Whether the twin's pre-trial recommendations hold up once they influence a real deployment is a question only the prospective design and deployment  can answer.

Second, although the architecture is temporally consistent, this constrains only the direction of information flow and does not by itself identify causal effects, so we cannot guarantee that the fitted twin recovers  \emph{true} causal effects of the intervention.
The fine-tuning trials are micro-randomized, but the action enters the denoiser non-additively, so our training may  not fully strip confounding from the generated trajectories (Appendix~\ref{sec:finetune}).
Recovering  causal effects may require importance-weighting the fine-tuning loss by the randomization probabilities.

\subsection*{Where this goes next}
We have demonstrated the framework on a single distribution shift within one program, so its reach beyond HeartSteps is a hypothesis to test rather than an established result.
We conjecture that the combination of strong generative priors, robust inference-time calibration, and a twin faithful enough to serve as an environment for pre-deployment algorithm design will transfer wherever a research team has small data from related trials, a written protocol for the new trial, and a domain scientist who can describe how the populations differ. 

\subsection*{Disclosure Statement}
The authors have no conflicts of interest to declare.

\subsection*{Contributions}
Z.X.: conceptualization, methodology, software, formal analysis, investigation, visualization, writing (original draft), and writing (review and editing).
Y.C.: methodology, software, investigation, and writing (review and editing).
C.N.: data curation and software.
N.S.: conceptualization and writing (review and editing).
A.H.G.: methodology, supervision, and writing (review and editing).
P.K.: resources, data curation, funding acquisition, and writing (review and editing).
E.E.: methodology, supervision, and funding acquisition.
S.M.: conceptualization, supervision, funding acquisition, and writing (review and editing).

\appendix

\renewcommand{\thefigure}{A\arabic{figure}}
\renewcommand{\thetable}{A\arabic{table}}
\setcounter{figure}{0}
\setcounter{table}{0}

\section{Exploratory Data Analysis of Raw Trajectories}
\label{appendix:eda}

Before introducing the digital twin, we summarize the hourly step-count data the twin must reproduce.
Table~\ref{tab:eda_summary} reports cohort sizes and step-count summary statistics for the three datasets used in this paper, all computed on the cleaned hourly transitions described in Section~\ref{sec:method}.

\begin{table}[h!]
\centering
\small
\caption{Raw step-count summary statistics per dataset (hourly observations after wear-time filtering).}
\label{tab:eda_summary}
\setlength{\tabcolsep}{4.5pt}
\begin{tabular}{l|rrrrrr}
\hline
Trial & participants & observation-hours & mean & median & std & \% zero-step hours \\
\hline
AoU (pre-pandemic) & 1{,}507 & 5{,}865{,}216 & 304 & 40  & 670 & 33.2\% \\
HS v2/v3           & 113     & 528{,}720     & 329 & 21  & 715 & 42.2\% \\
HS v4              & 72      & 621{,}428     & 350 & 129 & 628 & 11.5\% \\
\hline
\end{tabular}
\end{table}

Two patterns merit highlighting.
First, despite very different cohort sizes, the three trials have similar mean and standard deviation of hourly step counts ($\bar{s} \in [304, 350]$, $\sigma \in [628, 715]$), suggesting AoU is a viable pre-training source.
Second, the median and the fraction of zero-step hours diverge sharply: HS v4's median (129) is about $6\times$ that of HS v2/v3 (21) and AoU (40), and v4 has only
11.5\% zero hours versus $\geq 33\%$ for the other trials. This is
consistent with v4 participants wearing the tracker while active, a return-to-office post-pandemic pattern that motivates the calibration step (Section~\ref{sec:reward_tilt}).

\begin{figure}[h!]
\centering
\begin{minipage}{0.48\textwidth}\centering (a)
\includegraphics[width=\textwidth]{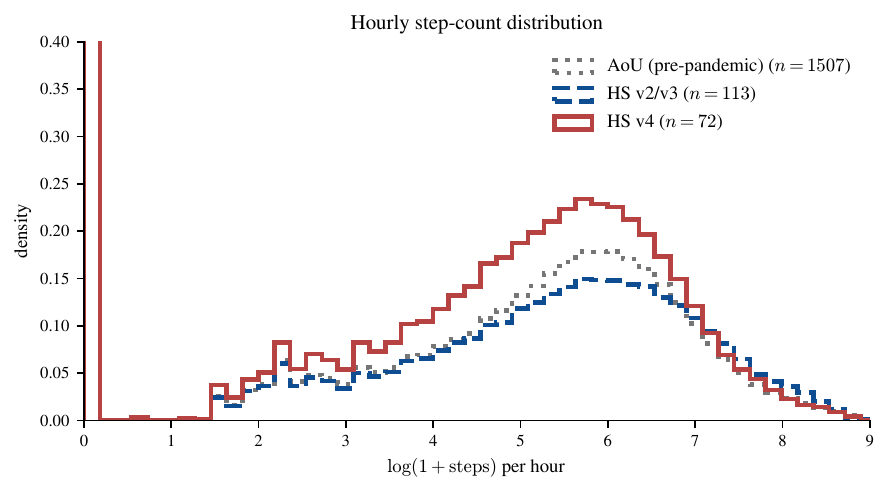}
\end{minipage}\hfill
\begin{minipage}{0.48\textwidth}\centering (b)
\includegraphics[width=\textwidth]{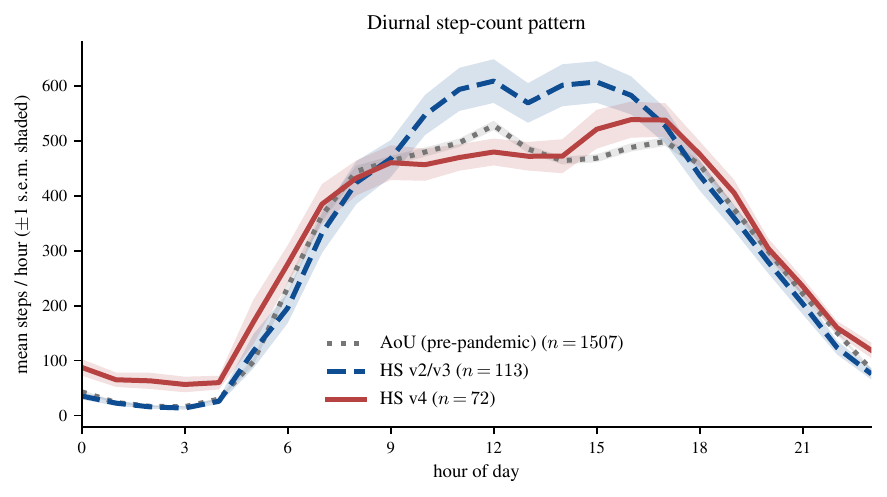}
\end{minipage}
\caption{(a) Distribution of $\log(1+\mathrm{steps})$ per hour across
the three trials. v4 shifts mass away from zero relative to v2/v3 and
AoU. (b) Mean step counts by hour-of-day (line), with a $\pm 1$ standard-error-of-the-mean band across participants (shaded). v2/v3 and AoU show a single mid-day peak. v4
exhibits the bimodal morning/evening pattern noted in
Section~\ref{sec:problem}.}
\label{fig:eda_distributions}
\end{figure}

\begin{figure}[h!]
\centering
\includegraphics[width=0.95\textwidth]{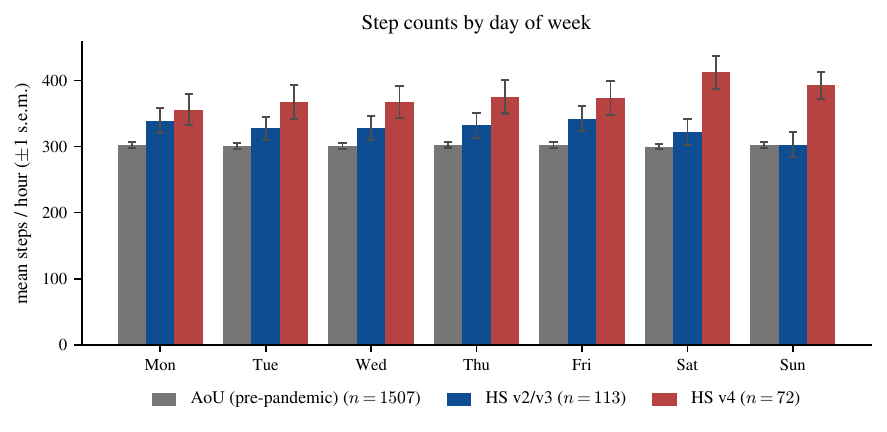}
\caption{Mean step counts per hour, by day-of-week, with $\pm 1$ standard-error-of-the-mean error bars across participants. The weekly cycle
is consistent across trials but with higher overall amplitude on v4.}
\label{fig:eda_dow}
\end{figure}

\begin{figure}[h!]
\centering
\includegraphics[width=0.95\textwidth]{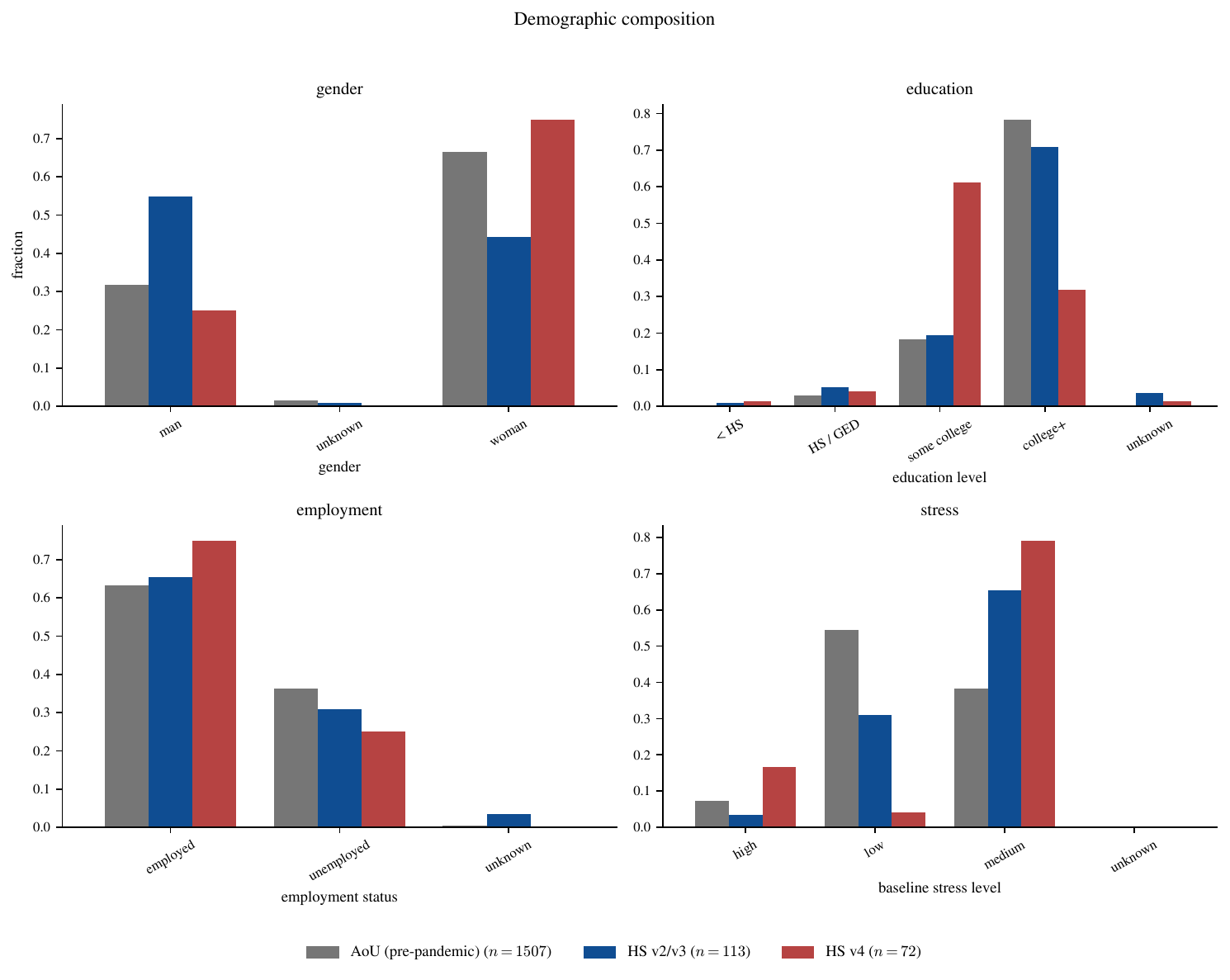}
\caption{Demographic composition of the three trials on four
conditioning fields used by the per-participant embedding: gender,
education, employment, and self-reported stress. Sample sizes are $n=1{,}507$ for \emph{All of Us} (pre-pandemic), $113$ for HS v2/v3, and $72$ for HS v4. 
}
\label{fig:eda_demo}
\end{figure}

\begin{figure}[h!]
\centering
\includegraphics[width=0.98\textwidth]{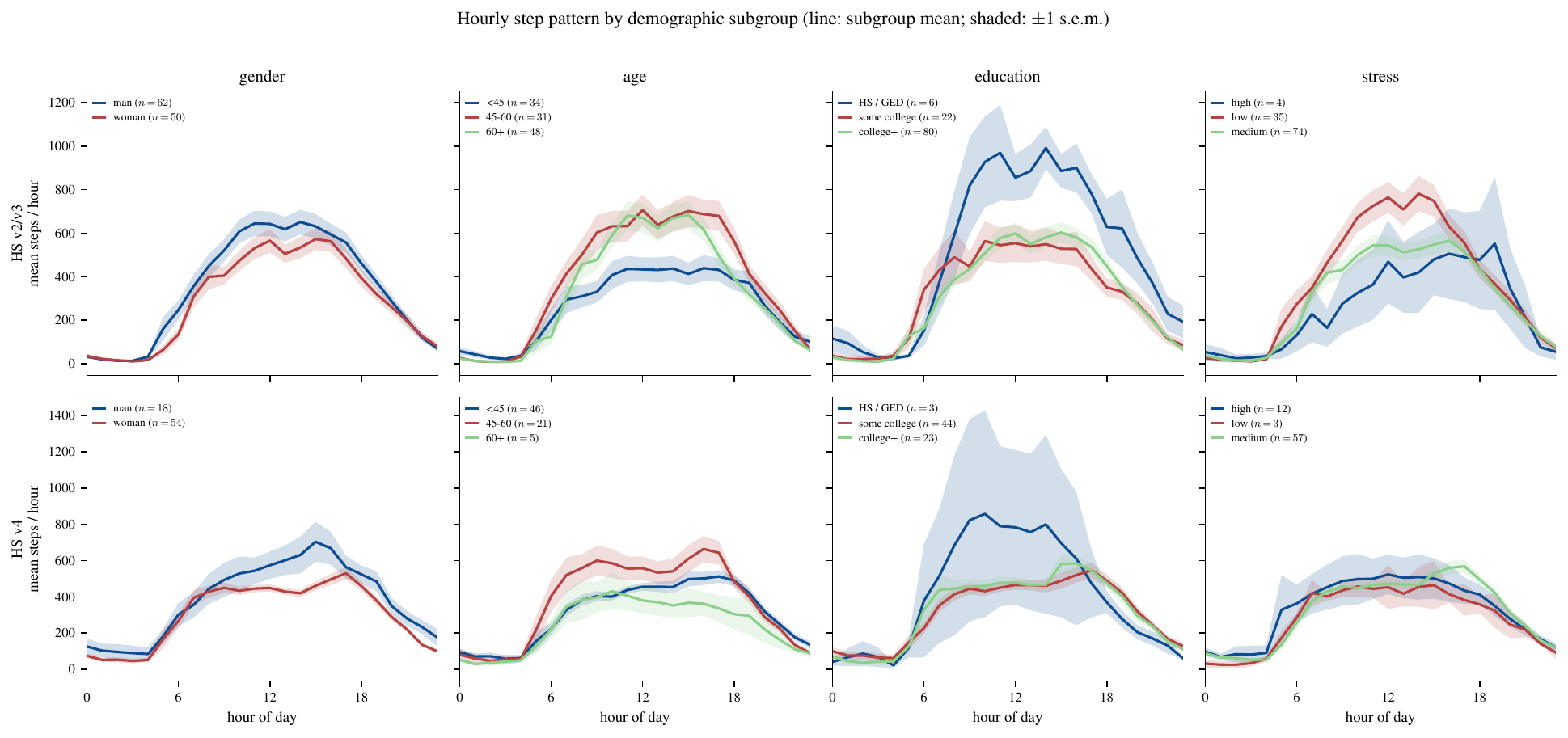}
\caption{Mean hourly step pattern stratified by demographics within
each HeartSteps trial (line: subgroup mean; shaded band: $\pm 1$ standard error of the mean across participants; subgroup $n$ in each legend). The trial-level pattern
(Figure~\ref{fig:eda_distributions}b) is preserved within every
subgroup, but subgroup amplitude and peak timing vary: this is the
heterogeneity that the per-participant conditioning embedding must
capture. Because the standard error scales as $1/\sqrt{n}$, the small v2/v3 subgroups carry visibly wide bands: the most extreme apparent contrasts, such as the elevated ``HS/GED'' education curve ($n=6$ in v2/v3, $n=3$ in v4), rest on a handful of participants and their bands overlap the other subgroups, so they should be read as within-noise rather than definitive. 
}
\label{fig:eda_conditioning}
\end{figure}

\begin{figure}[h!]
\centering
\includegraphics[width=0.98\textwidth]{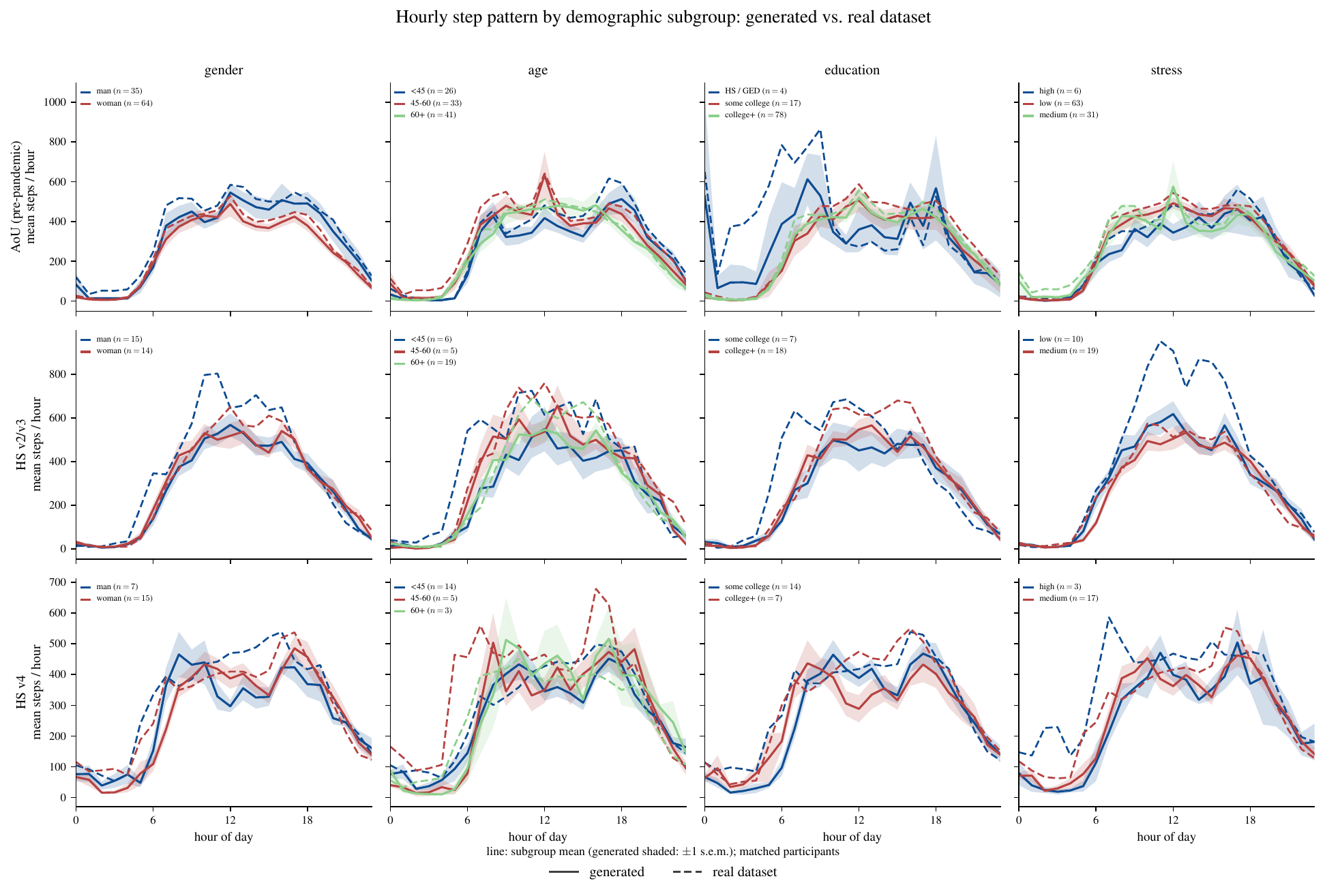}
\caption{Generated versus real hourly step pattern by demographic subgroup, across all three datasets.
Each row is one dataset (\emph{All of Us} pre-pandemic, HS v2/v3, HS v4); within every panel the generated subgroup mean (solid, with a $\pm 1$ standard-error-of-the-mean band across participants) is overlaid on the real subgroup mean (dashed) computed on the same matched participants; subgroup $n$ appears in each legend.
The generated curves track the real ones in level, peak timing, and subgroup ordering across gender, age, education, and stress.
Where they diverge it is in amplitude on the smallest strata rather than in direction, consistent with the magnitude-not-direction reading of Section~\ref{sec:demo}.
The v2/v3 and v4 rows use the calibrated twin; the \emph{All of Us} row uses the pre-trained model.}
\label{fig:eda_conditioning_gen}
\end{figure}

Figure~\ref{fig:eda_conditioning} shows the subgroup heterogeneity present in the real data.
Figure~\ref{fig:eda_conditioning_gen} shows that the twin's generated participants reproduce it.

\input{appendix_model}

\section{Participant Embedding Construction}
\label{appendix:embedding}

A key feature of a successful JITAI-Twin is its ability to capture participant heterogeneity, which is crucial for downstream online-algorithm design decisions such as how much to personalize across participants. In practice, we model participant heterogeneity by constructing a user embedding that represents each participant's latent traits. This embedding remains fixed throughout the trial, and all generation processes are conditioned on it.

We derive the user embeddings from baseline data collected through screening surveys administered before participants begin the trial. This baseline dataset is a structured collection of question-answer pairs, where each variable corresponds to a specific survey item. The dataset includes demographic information (e.g., age, gender) as well as measures such as mobile device usage frequency (see Table \ref{tab:baseline_app} for sample items).

\begin{table}[ht]
    \centering
    \caption{Sample questions and their valid values from the HeartSteps Baseline Data Dictionary}
    \begin{tabular}{p{6cm}|p{6cm}}
        Question & Valid Values \\
        \hline
        \hline
        What is your age? & 0-100 \\
        \hline
        How would you rate your comfort with using your mobile phone? & 1 (Not comfortable at all), \dots, 5 (Very comfortable) \\
        \hline
        ... & ...
    \end{tabular}
    \label{tab:baseline_app}
\end{table}

Two main challenges arise when constructing these embeddings.
First, different trials often use distinct survey instruments. To ensure the continual improvement of the JITAI-Twin, we must learn generalizable embeddings so that the generative model conditioned on embeddings from one trial can operate effectively on embeddings from another.
Second, a naive one-hot encoding of the baseline data can lead to a high-dimensional representation (348 dimensions for HeartSteps), which introduces computational challenges during both pre-training and fine-tuning. Moreover, one-hot encoding fails to capture the rich semantic information embedded in the question texts.

To address these challenges, we employ a frozen pre-trained T5 text encoder \citep{raffel2020t5} to generate user embeddings. Below, we describe the embedding construction procedure for our case study on HeartSteps using \emph{All of Us} data for pre-training. The procedure consists of three main steps.

\begin{itemize}
    \item \textbf{Questions Matching:} The \href{https://docs.google.com/spreadsheets/d/1Ey8MScRYZ9QyS4izVYScISLMb62QEhSM-ErbG27dNtw/edit?gid=1832128489&pli=1&authuser=0}{\emph{All of Us} baseline} dataset contains hundreds of questions, most of which are unrelated to those in the HeartSteps survey. Our first step is to identify overlapping or conceptually similar questions across the two datasets.
        \begin{enumerate}
            \item For each question in HeartSteps, we compute its word embedding and measure similarity with every question in \emph{All of Us}. The top ten most similar \emph{All of Us} questions are selected for further review. We then prompt a pre-trained LLM to evaluate whether each HeartSteps-\emph{All of Us} question pair is semantically relevant, along with a brief justification (see Figure \ref{fig:prompt} for a sample prompt).
            \item HeartSteps survey questions can be categorized into the following domains: DEM (demographics), PHONE/USE (mobile device usage), TRAIT (personality), EMPLOY (employment), LIFE (mental well-being), ROUT (daily routines), SELF (self-monitoring), SUPPORT (social support), and IPAQ (physical activity). Among these, we successfully matched questions in the DEM, EMPLOY, LIFE, and IPAQ categories. LIFE and IPAQ questions were direct matches, while DEM and EMPLOY questions required semantic matching due to differences in phrasing.
        \end{enumerate}
    \item \textbf{Constructing Text Input:} We summarize the matched question-answer pairs into a single compact text input, which removes redundancy and keeps the representation comparable across participants. We first compress the matched questions (see Table~\ref{tab:summary_questions} for the list of questions) and standardize the response formats. Notably, many IPAQ items are duplicates with varying units: for instance, ``How much time did you usually spend doing vigorous physical activities on one of those days (minutes per day)?'' and its counterpart in hours per day. We convert all such responses to minutes and compute the average across equivalent items.
    \item \textbf{Tag-based Prompting:} We further condense the key-value pairs obtained from the previous step into tag-based prompts following a rule-based method. Even though previous summarization greatly reduces redundancy, the final prompts can be further reduced by 1) removing repeated information in question-answer pairs (e.g., ``employed'' is self-evident for employment status), and 2) avoiding prompt dilation due to heterogeneous question length (e.g., IPAQ questions on average have six words compared to two for DEM and EMPLOY). 
    \begin{itemize}
        \item \textbf{Stress score} A summated Likert-type scale score is curated to quantify stress based on five categorical questions from the LIFE category with 4 for very stressful and 0 for totally non-stressful. We assign scores based on the categorical answers: Never (0), Almost Never (1), Sometimes (2), Fairly Often (3), and Very Often (4). An inverse scoring scheme is used based on the contexts of the questions. The stress score is calculated based on the arithmetic mean of individual questions. 
    \end{itemize}
    \item \textbf{Constructing User Embedding:} The summarized text input (see Figure~\ref{fig:text_input} for an example) is then processed by T5 to generate an embedding vector for each participant in the \emph{All of Us} dataset. We include additional ``empty heads'' to represent unmatched HeartSteps questions, enabling fine-tuning on HeartSteps data later.
\end{itemize} 

\begin{figure}[htbp]
\begin{promptbox}[Text input for T5]\small
{'Employment status': 'Employed',\\
 'Education level': 'College+',\\
 'Household size': 0,\\
 'Under-18 in household': 'Skipped',\\
 'Race': 'White',\\
 'Gender': 'Man',\\
 'Moderate activity avg. time (min/session)': 40.0,\\
 'Frequencies of moderate activity (last 7 days)': 2,\\
 'Walking avg. time (min/session)': 40.0,\\
 'Frequencies of walking (last 7 days)': 1,\\
 'Vigorous activity avg. time (min/session)': 35.0,\\
 'Frequencies of vigorous activity (last 7 days)': 3,\\
 'Sitting avg. time (min/day)': 622.5,\\
 'Can handle problems': 'Very Often',\\
 'Unable to handle problems': 'Never',\\
 'Feel nervous/stressed': 'Never',\\
 'Upset by things out of control': 'Never',\\
 'Have someone to turn to': 'Fairly Often'}
\end{promptbox}
\caption{Example of the final text input to T5 for one \emph{All of Us} participant, produced by the summarization and tag-based prompting steps.}
\label{fig:text_input}
\end{figure}

\begin{table}[ht]
\centering
\caption{Summary of selected questions}
\label{tab:summary_questions}
\begin{tabular}{|l|l|}
\hline
\textbf{Category} & \textbf{Variables} and \textbf{Types} \\ \hline
\textbf{DEM and EMPLOY} &
\begin{tabular}[c]{l|l}
Employment status & Categorical \\
Education level & Categorical \\
Household size & Count \\
Under-18 in household & Count \\
Race & Multi\_label \\
Gender & Multi\_label
\end{tabular} \\ \hline
\textbf{IPAQ} &
\begin{tabular}[c]{l|l}
Moderate activity average time (min/session) & Time Avg \\
Frequencies of moderate activity (last 7 days) & Count \\
Walking average time (min/session) & Time Avg \\
Frequencies of walking (last 7 days) & Count \\
Vigorous activity average time (min/session) & Time Avg \\
Frequencies of vigorous activity (last 7 days) & Count \\
Sitting average time (min/day) & Time Avg
\end{tabular} \\ \hline
\textbf{LIFE} &
\begin{tabular}[c]{l|l}
Can handle problems & Categorical \\
Unable to handle problems & Categorical \\
Feel nervous/stressed & Categorical \\
Upset by things out of control & Categorical \\
Have someone to turn to & Categorical
\end{tabular} \\ \hline
\end{tabular}
\end{table}

\begin{figure}[htbp]
\begin{promptbox}[Similarity-measurement prompt]\small
        You are an expert survey-harmonization assistant.\\
        Compare ONE HeartSteps question with ONE AoU question and decide if they\\
        Measure the same or highly similar construct. Be strict; consider answer\\
        Type and choices. If AoU field\_label is generic (e.g., Please specify)\\
        Infer intent using concept\_code/concept\_name/choices/form\_name.\\
        Return ONLY JSON with keys: 
        \begin{itemize}
            \item match (YES/NO)
            \item confidence (0..1), reason (short if YES)
            \item similarity\_type (one of: semantic\_paraphrase, same\_measure, same\_topic, subset, superset, metadata\_only)
            \item signals: \{type\_compatible: bool, choices\_overlap: 0..1, units\_match: bool\}
        \end{itemize}
\end{promptbox}
\caption{Prompt used to judge whether a HeartSteps--\emph{All of Us} question pair measures the same construct during question matching.}
\label{fig:prompt}
\end{figure}

\input{appendix_train_peft}

\section{Inference-time calibration: sampler derivation}
\label{appendix:calibration}

The reward-tilted sampling procedure used in Section~\ref{sec:reward_tilt} is implemented as a per-step modification to the standard DDPM reverse process.
We describe the linear-reward tilt used throughout the paper and the mean-shift target it is applied to.

\subsection*{Linear-reward tilt (used throughout the main paper).}
For a linear reward $r(\vs) = \langle \vv, \vs \rangle$ with $\vv \in \sR^{L}$ and tilt strength $\gamma$, where $\vs = \vx_0$ is the denoised trajectory of the reverse chain, the tilted reverse step is implemented by shifting the input to the denoising network and correcting the predicted noise: \begin{align*}
\text{shift}_t   &\leftarrow \frac{1 - \bar\alpha_t}{\sqrt{\bar\alpha_t}}\, (\gamma \vv) \\
\hat{\eps}_t     &\leftarrow \eps_\theta\bigl(\vx_t + \text{shift}_t,\,t,\, C_i\bigr)
                            - \frac{\sigma_t}{\sqrt{\bar\alpha_t}}\,(\gamma \vv),
\end{align*}
where $\bar\alpha_t$ and $\sigma_t = \sqrt{1 - \bar\alpha_t}$ are the standard VP-DDPM coefficients.
This is Algorithm 1 of \citet{moitra2026steering} specialized to a constant linear reward per hour, and integrated over the reverse chain it approximately shifts the mean of the generated sample by $\gamma \vv$ in the model's operating space.

\subsection*{Mean-shift calibration.} For the mean-shift target used
throughout Section~\ref{sec:reward_tilt} we set $\vv = \vmu^\star - \vmu^{\mathrm{base}}$, where $\vmu^\star$ is the target hourly mean in lin-log space and $\vmu^{\mathrm{base}}$ is the uncalibrated twin's hourly mean in the same space.
The lin-log transformation (see \citealp{narasimhan2024time}) is the model's native coordinate system.
The single calibration parameter $\gamma$ controls how aggressively to apply the shift.

\section{Comparison of calibration approaches}
\label{sec:calib_ablation}


Having introduced the calibration mechanism (Section~\ref{sec:reward_tilt}) and reported the calibrated pipeline in Table~\ref{tab:eval_main}, we now dissect how the calibration is achieved: which target source and which shifting mechanism buy the fidelity, and at what cost.
We compare two ways of applying a calibration target $\mu^\star_h$ (the desired per-hour mean of total steps, with $\mu^{\mathrm{base}}_h$ the mean of the uncalibrated twin).
The first is the reward-tilted sampler of Section~\ref{sec:reward_tilt}.
The second is a simpler post-hoc baseline, \emph{moment matching}: we take the twin's generated step counts and shift each hour additively,
\[
s'_{i,h} = s_{i,h} + \bigl(\mu^\star_h - \mu^{\mathrm{base}}_h\bigr),
\]
clipped at zero, so the per-hour mean matches the target by construction while the within-hour spread is left at the twin's own value.
Moment matching is given exactly the information calibration uses (the target per-hour \emph{mean}).

Table~\ref{tab:calibration} reports all of the calibration results. We read the v2/v3$\to$v4 hand-off (panel (b)) first and then the earlier AoU$\to$v2/v3 hand-off (panel (a)) the same way. We begin with the \emph{true} v4 hourly mean as the target (panel (b)'s top three rows) and turn to the forecast targets produced by the method of Section~\ref{sec:llm} (its bottom two rows).

\begin{table}[h!]
\centering
\footnotesize
\setlength{\tabcolsep}{2pt}
\caption{Calibration at both pretend-prospective hand-offs. \textbf{(a)} Calibrating the AoU-pretrain twin toward held-out v2/v3; \textbf{(b)} calibrating the v2/v3 twin toward held-out v4.\protect\footnotemark}
\label{tab:calibration}
\newcommand{\pmse}[1]{\ensuremath{{\scriptstyle\,(\pm #1)}}}

\textbf{(a) AoU-pretrain twin $\to$ held-out v2/v3}\\[3pt]
\resizebox{\textwidth}{!}{%
\begin{tabular}{l|ccccccc}
\toprule
Source of target & Daily W (log) $\downarrow$ & ACF-MAE (log) $\downarrow$
  & Run-len.\ W $\downarrow$ & h-mean MAE $\downarrow$
  & bu-acf $\to 1$ & bu-run $\to 1$ & v2/v3 data? \\
\midrule
none (uncalibrated)             & 10.82\pmse{0.22} & 0.274\pmse{0.032} & 2.06\pmse{0.28} & 181\pmse{39} & 0.36\pmse{0.11} & 0.24\pmse{0.09} & no \\
true v2/v3, moment matching  & 10.45\pmse{0.23} & \textbf{0.074}\pmse{0.023} & 4.33\pmse{0.29} & \textbf{48}\pmse{22} & 0.28\pmse{0.08} & \textbf{0.50}\pmse{0.19} & yes \\
true v2/v3, calibration         & \textbf{10.05}\pmse{0.25} & 0.223\pmse{0.034} & \textbf{1.49}\pmse{0.30} & 67\pmse{25} & \textbf{0.43}\pmse{0.12} & 0.35\pmse{0.13} & yes \\
AoU marginal, calibration       & 10.25\pmse{0.27} & 0.243\pmse{0.034} & 1.68\pmse{0.28} & 114\pmse{34} & 0.40\pmse{0.11} & 0.29\pmse{0.11} & no \\
scientist + LLM, calibration    & 10.22\pmse{0.26} & 0.241\pmse{0.034} & 1.66\pmse{0.28} & 106\pmse{33} & 0.40\pmse{0.11} & 0.30\pmse{0.11} & no \\
\bottomrule
\end{tabular}%
}

\vspace{8pt}

\textbf{(b) v2/v3 twin $\to$ held-out v4}\\[3pt]
\resizebox{\textwidth}{!}{%
\begin{tabular}{l|ccccccc}
\toprule
Source of target & Daily W (log) $\downarrow$ & ACF-MAE (log) $\downarrow$
  & Run-len.\ W $\downarrow$ & h-mean MAE $\downarrow$
  & bu-acf $\to 1$ & bu-run $\to 1$ & v4 data? \\
\midrule
none (uncalibrated)          & 8.43\pmse{0.19} & 0.081\pmse{0.025} & 0.74\pmse{0.37} & 49\pmse{22} & 0.35\pmse{0.09} & 0.46\pmse{0.14} & no \\
true v4, moment matching  & 9.96\pmse{0.22} & \textbf{0.069}\pmse{0.022} & 2.64\pmse{0.49} & \textbf{36}\pmse{16} & 0.33\pmse{0.08} & \textbf{0.72}\pmse{0.16} & yes \\
true v4, calibration         & \textbf{8.32}\pmse{0.17} & 0.107\pmse{0.025} & \textbf{0.30}\pmse{0.22} & 48\pmse{15} & \textbf{0.35}\pmse{0.09} & 0.58\pmse{0.14} & yes \\
v23 marginal, calibration    & 8.33\pmse{0.17} & 0.116\pmse{0.025} & 0.39\pmse{0.22} & 70\pmse{15} & 0.33\pmse{0.08} & 0.57\pmse{0.13} & no \\
scientist + LLM, calibration & 8.34\pmse{0.18} & 0.108\pmse{0.025} & 0.30\pmse{0.22} & 53\pmse{16} & 0.33\pmse{0.08} & 0.57\pmse{0.13} & no \\
\bottomrule
\end{tabular}%
}
\end{table}
\footnotetext{The last two columns (bu-acf, bu-run) are closer-to-1-better, so the bold entry is nearest 1; the remaining columns are lower-is-better. The daily W column is the daily-profile Wasserstein of Table~\ref{tab:eval_main}, computed under the same protocol and per-panel $N$ ($500$ in (a), $150$ in (b)). The remaining columns are computed on a held-out test split of the target deployment with per-participant matching (the protocol of Table~\ref{tab:credit}'s non-distributional columns), so they are comparable within this table but not numerically identical to the footprint-matched columns of Table~\ref{tab:eval_main}. More details are in Appendix~\ref{appendix:metrics_baselines}.}

Both procedures pull the base toward v4's per-hour level, but they buy that fidelity differently.
Moment matching hits v4's level by construction (lowest per-hour-of-day mean MAE, 36), but shifting every hour by a constant and clipping at zero fuses the twin's near-zero overnight hours into long bouts: its run-length Wasserstein of 2.64 is worse than even the \emph{uncalibrated} twin's 0.74, and the same damage shows up in the joint daily profile, where it moves the twin \emph{away} from v4 (daily Wasserstein $8.43 \to 9.96$), so the post-hoc shift makes whole-day realism worse than doing nothing.
Calibration instead moves the mean from inside the generative process, so the zero spike and bout structure survive (run-length $0.30$, the best in the panel) while the daily-profile fit improves ($8.43 \to 8.32$, matching the true-target calibration).
It thus preserves the daily profile and wins the activity-bout structure, conceding the per-hour mean (and the short-lag autocorrelation) to moment matching's exact per-hour reshaping.

On between-participant heterogeneity the two are close, and neither manufactures spread the base twin lacks (the base under-disperses at bu-acf $0.35$, bu-run $0.46$ against a target of $1.0$).
Moment matching's higher bu-run ($0.72$ versus calibration's $0.58$) is not a real gain: it is spread in the unrealistic bouts its clipping creates, whereas calibration preserves the spread while keeping each participant's run lengths realistic.

\begin{figure}[h!]
\centering
\begin{subfigure}{0.49\textwidth}\centering
\includegraphics[width=\textwidth,height=0.72\textwidth,keepaspectratio]{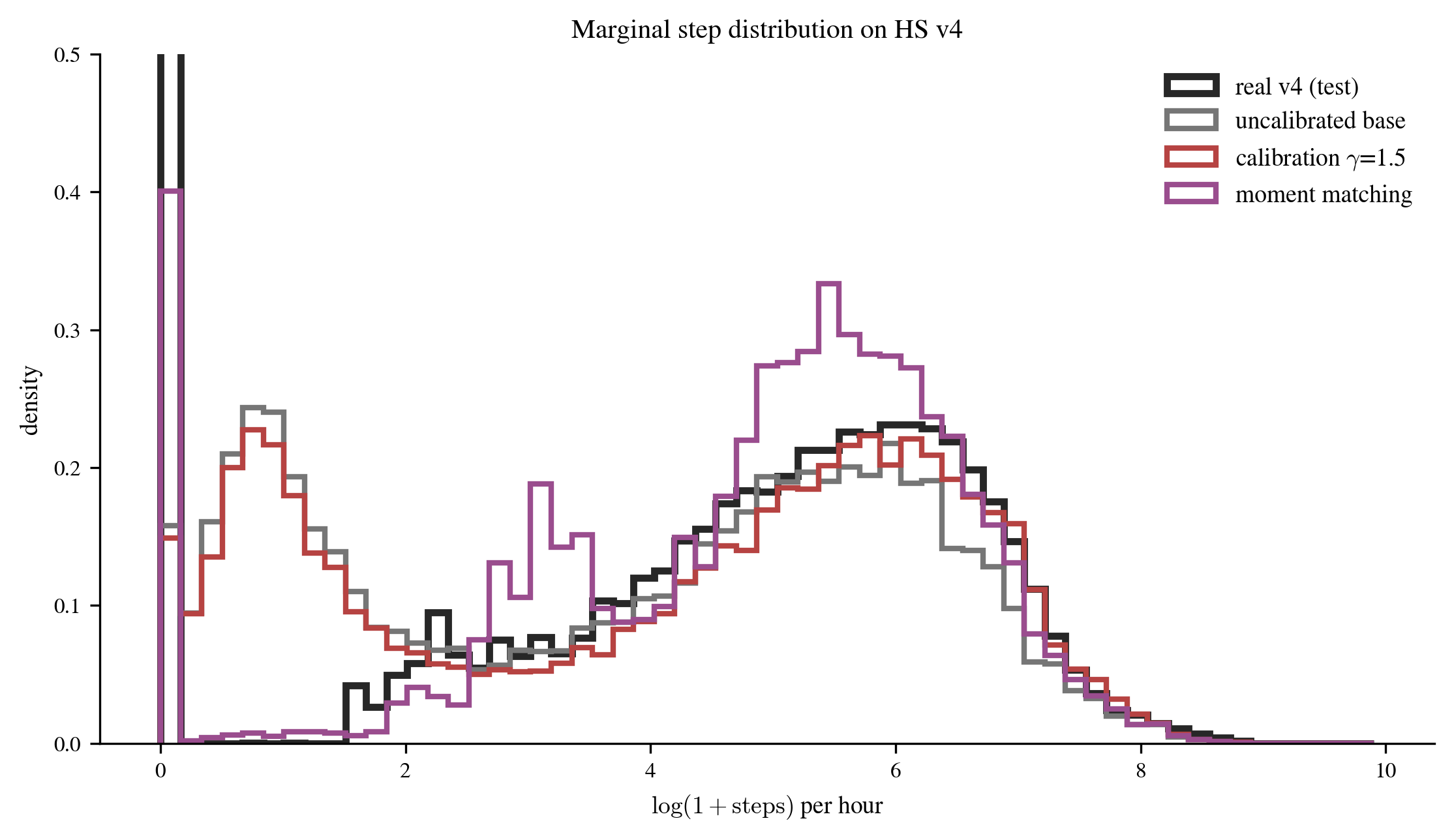}
\caption{Marginal step distribution.}
\label{fig:step_hist_pretrain}
\end{subfigure}\hfill
\begin{subfigure}{0.49\textwidth}\centering
\includegraphics[width=\textwidth,height=0.72\textwidth,keepaspectratio]{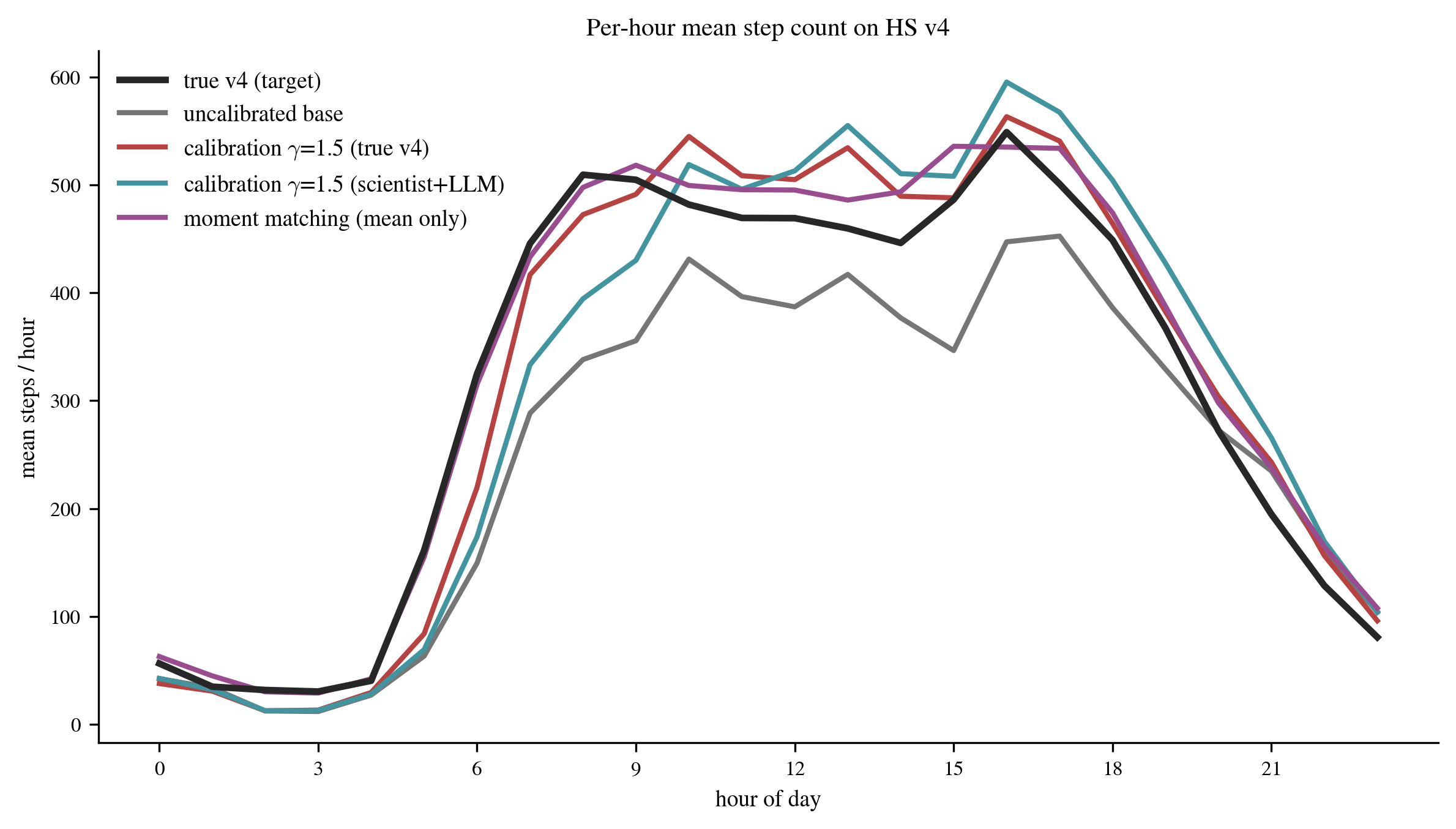}
\caption{Per-hour mean vs.\ true v4 target.}
\label{fig:hourly_mean_v4}
\end{subfigure}
\caption{Calibration on HS v4.
(a)~Marginal $\log(1+\text{steps})$ distribution: the uncalibrated base already matches real v4 on the zero spike and the upper tail.
(b)~Per-hour mean, shown against the true v4 target (black, the same target used in the main-text figure): in-process calibration ($\gamma=1.5$; red and teal) tracks the target's bimodal shape, whereas the post-hoc moment-matching baseline (violet) mostly rescales the flatter base profile.}
\label{fig:calib_marginal}
\end{figure}

Figure~\ref{fig:hourly_mean_v4} visualizes the per-hour mean curve: both procedures pull the uncalibrated twin's curve up toward the v4 reference, but calibration tracks v4's bimodal morning--afternoon shape while the post-hoc shift only rescales the twin's flatter profile.

\section{Producing the calibration target: LLM prompts and coordinate conversion}
\label{appendix:llm}

The calibration target is produced in two steps with GPT-5.4-mini (Azure OpenAI); the exact prompts are reproduced in Figures~\ref{fig:llm_prompt_step1} and~\ref{fig:llm_prompt_step2}.
The prompt never names the target deployment or the program and uses no post-hoc result. It supplies only pre-deployment knowables.
Because the API is not deterministic, we draw $N=8$ independent two-step chains and use the per-hour median of their normalized shapes, scaled by the prior deployment's daily total, as the target $\mu^\star$. The shape is stable across draws.

\subsection{Step 1: qualitative reasoning (protocol-only prompt)}

\begin{figure}[htbp]
\begin{promptbox}[Step 1 prompt]\small
You are a behavioral-science researcher who has worked on a long-running series of physical-activity nudge micro-randomized trials. We are about to launch the \emph{next} deployment of this program in a new city and need to anticipate, before any data are collected, how its participants' daily step pattern will differ from the cohorts we have already studied. Use only the following:
\begin{itemize}[leftmargin=1.4em,topsep=2pt,itemsep=1pt]
\item two prior intervention trials of this program, both in Seattle, enrolling adults with coronary artery disease aged $\ge 40$ recruited from a phase~II cardiac-rehabilitation program;
\item the upcoming deployment in greater Los Angeles, enrolling sedentary adults with body-mass index 25--45, aged 18--65.
\end{itemize}
Reason from first principles about how the location and eligibility criteria would shift the \emph{hourly} step-count distribution within a typical day, relative to the prior trials. Structure your answer as: (1)~direction of overall daily step count; (2)~diurnal pattern; (3)~within-hour variability; (4)~one-sentence summary. Be specific about hours of day; under {\raise.17ex\hbox{$\scriptstyle\sim$}}500 words.
\end{promptbox}
\caption{Step 1 prompt (qualitative reasoning) as run in the HeartSteps replay, instantiating the generic template of Figure~\ref{fig:generic_calib_prompt}.}
\label{fig:llm_prompt_step1}
\end{figure}

A representative response is summarized in Section~\ref{sec:llm}. Full outputs are provided in the supplementary material.

\subsection{Step 2: diurnal shape (level from data)}

\begin{figure}[htbp]
\begin{promptbox}[Step 2 prompt]\small
You previously predicted how the \emph{next} deployment's step pattern will differ from the prior trials: [\emph{Step 1 response verbatim}]. Here are the observed prior-trial statistics (raw hourly step counts, pooled across all prior-trial participants and days), for reference: [\emph{24-hour table of prior-trial mean, median, std}]. Output the predicted diurnal \emph{shape}: 24 non-negative numbers (hours 0--23) giving the fraction of the day's steps taken in each hour, summing to one. Commit to where activity sits based on your reasoning; do not assume the new shape matches the prior cohort. Output format (required, strict): exactly one JSON object on its own line, no commentary, no markdown fence, in this shape: \texttt{\{"shape": [s0, ..., s23]\}}.
\end{promptbox}
\caption{Step 2 prompt (numeric target) as run in the HeartSteps replay, instantiating the generic template of Figure~\ref{fig:generic_calib_prompt}.}
\label{fig:llm_prompt_step2}
\end{figure}

\textbf{Why shape only.} An earlier version asked the model to translate its prediction into hourly \emph{means} given the prior-trial table. Anchored on those numbers, the model returned the prior-trial shape with small per-hour deviations, mislocating v4's activity to the afternoon. Eliciting the normalized shape removes that anchor, and the daily \emph{total} is more reliably supplied from the prior trial (which estimates v4's level to within a few percent) than predicted by the model (which systematically under-predicted it for this sedentary cohort).

\textbf{Validation.} The resulting target matches v4's level (scaled total $7{,}712$ vs.\ real v4's ${\approx}\,7{,}500$) and recovers its morning-shifted shape. Calibrating toward it reaches a per-hour-of-day MAE of $53$ at $\gamma=1.5$, close to the oracle target's $48$ and far below the prior-shape (v23-marginal) target's $70$ (Section~\ref{sec:llm}).
We take the per-hour median of the normalized shape over $N=8$ draws, scale it by the prior-trial daily total, and tile it to a 168-hour (day-of-week $\times$ hour-of-day) vector to form $\mu^\star$ for the calibration sampler.

\subsection{Raw $\to$ lin-log calibration}
\label{appendix:llm_calib}

The LLM's prediction is in raw step coordinates. The diffusion model operates in lin-log coordinates (a piecewise transform whose log branch tames the heavy right tail of step counts).
The two coordinate systems differ not just by a monotone function but \emph{also} by the order in which averaging and transformation are applied: in raw space the practitioner most naturally specifies $\E[X]$, while the diffusion model's natural target is $\E[\mathrm{lin\_log}(X)]$, and these differ in general because $\mathrm{lin\_log}$ is concave on the upper tail and the underlying distribution is heavy-tailed.

We estimate the per-hour conversion factor $f_h = \E[\mathrm{lin\_log}(X)]_h\,/\,\mathrm{lin\_log}(\E[X])_h$ from the v23 training data and apply it to both the LLM-supplied target and the uncalibrated twin's baseline mean before computing the calibration vector $\vv = \vmu^\star - \vmu^{\mathrm{base}}$.
Empirically $f_h \in [0.50,\,1.0]$ with mean 0.71, so omitting the conversion biases the calibration target up by about $1/0.71 \approx 1.4\times$ and produces over-calibration (which we observed when running without the conversion, Section~\ref{sec:llm}).

\section{Evaluation metrics and baseline simulators}
\label{appendix:metrics_baselines}

\subsection{Metric definitions}

All metrics for Table~\ref{tab:eval_main} are computed at the cohort level against the full participant set of the target deployment.
The daily-profile Wasserstein carries a strong sample-size-dependent bias, so rows scored on unequal amounts of data are not comparable on it: a row emitting the full cohort's hours would hold an unearned advantage over a row emitting fewer.
This metric is therefore scored under a \emph{footprint-matched} protocol: each row's cohort is cut into disjoint 30-participant folds covering every participant (wrap-filling the last fold), each participant contributes one contiguous four-week block, each fold is compared against the full real target cohort, and the point estimate averages the folds over several random fold assignments with a fixed seed.
The remaining four columns are far less sample-size sensitive and are computed once on each row's full output against the full target cohort.
Standard errors are participant bootstraps: the participants are resampled with replacement (on both sides for the full-output columns) and the estimate is re-run on each replicate.
The target's own reference row of Table~\ref{tab:eval_main} is a split-half comparison on these columns: the target cohort is split at random into two halves, one half scored against the other, averaged over random splits.
We index the $N$ participants by $i$ and write each participant's hourly step series over the $L = 168$ hours of a week as $S^{(i)} = (S^{(i)}_1, \dots, S^{(i)}_L) \in \sR_{\ge 0}^{L}$, using $S^{(i)}$ for the real series and $\tilde S^{(i)}$ for the generated one, and $\mathrm{SD}_i[\cdot]$ for the standard deviation taken across participants.

\subsection*{Daily-profile Wasserstein distance (distributional fit).}
We compare the distribution of whole daily activity profiles rather than of pooled hourly values.
For every participant-day with all 24 hours observed, form the day vector $D = \bigl(\log(1+S_{1}), \dots, \log(1+S_{24})\bigr) \in \sR^{24}$; days with zero steps in every hour are non-wear artifacts and are excluded from both the generated and the real side.
Writing $\hat P_{\mathrm{gen}}$ and $\hat P_{\mathrm{real}}$ for the empirical distributions of the generated and real day vectors, the metric is the first Wasserstein distance with Euclidean ground metric,
\[
  W_1\bigl(\hat P_{\mathrm{gen}}, \hat P_{\mathrm{real}}\bigr)
  = \min_{\pi \in \Pi(\hat P_{\mathrm{gen}}, \hat P_{\mathrm{real}})} \E_{(X,Y) \sim \pi} \bigl\| X - Y \bigr\|_2 ,
\]
where $\Pi$ is the set of couplings of the two empirical distributions.
We compute it exactly: both sides are subsampled to a common size $N$ with uniform weights, where the optimal coupling is a permutation (Birkhoff--von Neumann), so the distance reduces to an optimal assignment problem solved exactly by the Hungarian algorithm, with no entropic regularization and no projection approximation.
The empirical multivariate Wasserstein distance carries an upward finite-sample bias that grows as $N$ shrinks, so $N$ is held fixed within each panel at the largest value every row can supply per fold ($N = 500$ for the v2/v3 hand-off and $N = 150$ for v4, whose wear-time mask leaves fewer fully observed 24-hour participant-days); entries are therefore compared only within a panel, and the split-half floor gives the irreducible value at that $N$.
Earlier drafts reported the one-dimensional Wasserstein between the pooled hourly $\log(1+\text{steps})$ distributions, treating every participant-hour as one draw.
That distance scores the time-marginalized step distribution alone and is blind to temporal and between-participant structure: a resampler of a matching source population reproduces the pooled marginal essentially for free (PSRS reached $0.13$ at the first hand-off, at the floors), while the joint daily profile still separates the simulators, so every table now reports the daily-profile distance instead.

\subsection*{Short-lag autocorrelation MAE}
For a series $x \in \sR^{L}$ with mean $\bar x$, the lag-$k$ sample autocorrelation is
\[
  \rho_k(x) = \frac{\sum_{t=1}^{L-k}(x_t - \bar x)(x_{t+k} - \bar x)}{\sum_{t=1}^{L}(x_t - \bar x)^2} .
\]
Averaging each lag over participants, $\bar\rho_k^{\mathrm{gen}} = \tfrac1N \sum_i \rho_k(\tilde S^{(i)})$ and $\bar\rho_k^{\mathrm{real}} = \tfrac1N \sum_i \rho_k(S^{(i)})$, the metric is the mean absolute error over the first six lags,
\[
  \mathrm{ACF\text{-}MAE} = \frac{1}{6}\sum_{k=1}^{6} \bigl| \bar\rho_k^{\mathrm{gen}} - \bar\rho_k^{\mathrm{real}} \bigr| .
\]
Lower values mean the generated within-day temporal dependence matches the real one.

\subsection*{Active-period run-length Wasserstein.}
With the HeartSteps active-hour threshold $\tau = 100$ steps, let $a^{(i)}_t = \mathbbm{1}[\,S^{(i)}_t \ge \tau\,]$ be the active indicator and define a \emph{run} as a maximal block of consecutive active hours.
Let $\mathcal{R}_{\mathrm{real}}$ and $\mathcal{R}_{\mathrm{gen}}$ be the multisets of run lengths pooled over all participants for the real and generated data.
The metric is $W_1$ between their empirical distributions, $W_1(\hat F_{\mathcal{R}_{\mathrm{gen}}}, \hat F_{\mathcal{R}_{\mathrm{real}}})$.
A simulator with the right marginal but the wrong stitching of hours into bouts scores poorly.

\paragraph{Threshold robustness ($\tau = 100$ vs.\ $250$ steps/hour).}
The $\tau = 100$ activity threshold follows the HeartSteps active-hour convention.
Because that cutoff is somewhat arbitrary, we recompute the run-length Wasserstein at a stricter $\tau = 250$ steps/hour for every Table~\ref{tab:eval_main} candidate, under the same protocol and bootstrap standard errors as Table~\ref{tab:eval_main}'s run-length column (Table~\ref{tab:run_length_threshold}); the $\tau=100$ column repeats that column of Table~\ref{tab:eval_main}.
The two thresholds agree on the calibrated pipeline (run-length W $0.22$ vs.\ $0.20$ at the v4 hand-off, the two thresholds within rounding of each other).
At the stricter threshold the PSRS and parametric baselines score lower on this single metric, because cutting at $250$ steps/hour discards much of the low-activity structure where they diverge from real activity, while KNN worsens at the first hand-off; the between-participant heterogeneity those baselines erase (Table~\ref{tab:eval_main}) is unaffected by the threshold.
\input{Figure/eval/run_length_threshold_table}

\subsection*{Between-participant heterogeneity ratios (bu-acf, bu-run).}
For a per-participant statistic $\phi : \sR^{L} \to \sR$, the heterogeneity ratio compares the across-participant spread of the generated and real populations,
\[
  \mathrm{HR}(\phi) = \frac{\mathrm{SD}_i\bigl[\phi(\tilde S^{(i)})\bigr]}{\mathrm{SD}_i\bigl[\phi(S^{(i)})\bigr]} ,
\]
so $\mathrm{HR} = 1$ means the synthetic population varies between individuals exactly as much as v4 does, and $\mathrm{HR} < 1$ means the synthetic participants are more alike than the real ones.
We report it for the two within-participant temporal statistics: $\phi_{\mathrm{acf}}(x) = \tfrac16\sum_{k=1}^6 \rho_k(x)$ gives \textbf{bu-acf}, and $\phi_{\mathrm{run}}(x)$, the mean length of $x$'s active runs, gives \textbf{bu-run}.
A simulator that imposes one shared temporal dynamic on every participant drives these ratios toward zero, whereas a twin that draws each participant's dynamics from their own conditional distribution can keep them near one.

For completeness we also track the heterogeneity ratio of the per-participant mean and total step count, the per-participant peak activity hour, and the per-participant active-hour fraction, along with several other diagnostics (Jensen-Shannon divergence, per-hour-of-day median MAE, day-to-day consistency, and DTW on each participant's mean-day curve). All are reported in the supplementary material but elided from the main text to keep the comparison focused. The per-hour-of-day mean MAE (h-mean MAE), which the calibration target directly optimizes, is surfaced in the calibration comparison of Table~\ref{tab:calibration}.

\subsection*{Scoring scale and inactivity flooring.} 
The daily-profile Wasserstein, ACF-MAE, and bu-acf are computed on the $\log(1+\text{steps})$ scale, which avoids over-crediting exact reproduction of the zero and low-count hours that dominate the raw distribution; run-length W and bu-run are threshold-based (active hour $\ge 100$ steps) and stay on the raw scale.
For the v2/v3 hand-off (Table~\ref{tab:eval_main}a), where about $42\%$ of hours are inactive, both generated and real step counts are floored to zero for $\log(1+\text{steps}) \le 3$ (about $19$ steps/hr), a symmetric inactivity threshold; v4's lower zero rate is already matched, so no floor is applied in panel (b).
In the v2/v3 hand-off the calibration gain in Table~\ref{tab:calibration}(a) is most visible in the per-hour mean (h-mean MAE) and the daily-profile Wasserstein ($10.82 \to 10.05$ toward the true target).

\subsection{Baseline simulators}

The main text reports three representative baseline simulators (KNN, per-state rejection sampling (PSRS), and linear SEM); for completeness we describe all four here, adding the GRU sequence generator, the strongest neural variant of the transition family.
In contrast to the diffusion twin, none of them use the baseline survey: each is a \emph{state-action transition model} that samples the next hourly step count conditional on the current state, namely the previous step count and the clock, together with the action.
Writing $\mathrm{dow}(t)$ and $\mathrm{hod}(t)$ for the day-of-week and hour-of-day at hour $t$, and $a_t \in \{0,1\}$ for the intervention delivered at $t$, every baseline samples from
\begin{equation}
  \mathrm{steps}_t \ \sim \ f\big(\mathrm{steps}_{t-1},\ \mathrm{dow}(t),\ \mathrm{hod}(t),\ a_t\big).
  \label{eq:baseline_transition}
\end{equation}
At each hand-off we fit each model on the pooled one-step transitions of the source data (\emph{All of Us} at the first hand-off, where the action column is identically zero because no interventions were delivered; v2/v3 at the second) and generate by replaying each target participant's \emph{real} clock-and-action sequence, rolling the model forward \emph{closed-loop}: the model's own sampled step is fed back as the next $\mathrm{steps}_{t-1}$, so sampling error and any learned dynamics propagate exactly as they would when the twin drives an online algorithm.
We use a common notation across the four simulators.
Write $y_t = \log(1+\mathrm{steps}_t)$ for the log-step in hour $t$, with day-of-week $d_t$, hour-of-day $h_t$, and action $a_t\in\{0,1\}$.
At hour $t$ the previous step $y_{t-1}$ is known and the \emph{current} step $y_t$ is what the simulator generates.
The simulators are fit on the pooled logged transitions of the source data $\mathcal{D}=\{(y_j, d_j, h_j, a_j, y^{\mathrm{next}}_j)\}_{j=1}^{N}$, where transition $j$ records a step $y_j$ observed at day-of-week $d_j$ and hour $h_j$ under action $a_j$, together with its realized next step $y^{\mathrm{next}}_j$.
Let $\mathrm{onehot}(\cdot)$ be a one-hot encoding ($\mathrm{onehot}(d)\in\{0,1\}^7$, $\mathrm{onehot}(h)\in\{0,1\}^{24}$), $\tilde{y}=(y-\bar{y})/s_y$ a standardized step, and $\phi_t=\big(y_{t-1},\, \mathrm{onehot}(d_t),\, \mathrm{onehot}(h_t),\, a_t\big)\in\mathbb{R}^{33}$ the parametric feature, stacked into the design matrix $\Phi\in\mathbb{R}^{N\times 33}$ with target vector $\mathbf{y}=(y^{\mathrm{next}}_j)_{j=1}^{N}$.
Generation replays each held-out participant's $(d_t,h_t,a_t)$ closed-loop, feeds the sampled $y_t$ back as the next $y_{t-1}$, and recovers $\mathrm{steps}_t=\exp(y_t)-1$ (clipped to $[0,e^{12}]$).

The two \emph{retrieval} samplers (KNN and PSRS; Algorithms~\ref{alg:knn} and~\ref{alg:rejection}) return a logged next-step $y^{\mathrm{next}}_j$; the PSRS variant follows \citet{tang2022towards}.
In Tang et al.'s per-state rejection sampler a logged transition is accepted with probability proportional to the importance weight $\pi(a\mid s)/\pi_b(a\mid s)$, where $\pi_b(a\mid s)$ is the logging (randomization) propensity with which that action was actually delivered.
Because we replay each participant's actually delivered action sequence, the target $\pi(\cdot\mid s)$ is a point mass on the delivered action, so the acceptance reduces to keeping the transitions whose logged action matches the replayed one: the propensity cancels exactly, and a rarely delivered nudge simply leaves few candidates.
The two \emph{parametric} models (linear SEM and GRU; Algorithms~\ref{alg:linsem} and~\ref{alg:gru}) draw $y_t$ from a Gaussian in $\phi_t$.

\begin{algorithm}[h!]
\caption{KNN retrieval simulator ($K=10$).}
\label{alg:knn}
\begin{algorithmic}[1]
\Require query state $s_t=(y_{t-1},d_t,h_t)$ and action $a_t$; logged transitions $\mathcal{D}=\{(y_j, d_j, h_j, a_j, y^{\mathrm{next}}_j)\}$
\State $\mathcal{A} \gets \{\, j : a_j = a_t \,\}$
\State $z \gets (\tilde{y}_{t-1}, \mathrm{onehot}(d_t), \mathrm{onehot}(h_t))$, \quad $z_j \gets (\tilde{y}_j, \mathrm{onehot}(d_j), \mathrm{onehot}(h_j))$ \Comment{$\tilde{y}$: standardized $y$}
\State $\mathcal{N} \gets \operatorname*{arg\,min}\limits^{K}_{\,j \in \mathcal{A}}\, \lVert z - z_j \rVert_2$
\State $w_j \gets (\lVert z - z_j \rVert_2 + \epsilon)^{-1} \big/ \textstyle\sum_{k\in\mathcal{N}}(\lVert z - z_k \rVert_2 + \epsilon)^{-1}$
\State \Return $y_t \gets \sum_{j\in\mathcal{N}} w_j\, y^{\mathrm{next}}_j$
\end{algorithmic}
\end{algorithm}

\begin{algorithm}[h!]
\caption{Per-state rejection sampling (PSRS) retrieval simulator (Tang et al.; bandwidth $b=0.75$).}
\label{alg:rejection}
\begin{algorithmic}[1]
\Require query $(y_{t-1},d_t,h_t,a_t)$; logged transitions $\mathcal{D}=\{(y_j, d_j, h_j, a_j, y^{\mathrm{next}}_j)\}$
\State $\mathcal{R} \gets \{\, j : a_j = a_t,\ d_j = d_t,\ h_j = h_t \,\}$, backing off $d_j$ then $h_j$ until nonempty
\State sample $j \sim \rho$ with $\rho_j \propto \exp\!\big(-\tfrac{1}{2}\big((y_{t-1} - y_j)/b\big)^2\big)$, \ $j \in \mathcal{R}$
\State \Return $y_t \gets y^{\mathrm{next}}_j$
\end{algorithmic}
\end{algorithm}

\begin{algorithm}[h!]
\caption{Linear SEM parametric simulator (ridge $\lambda=1$).}
\label{alg:linsem}
\begin{algorithmic}[1]
\Require design matrix $\Phi$ and targets $\mathbf{y}$; query feature $\phi_t$
\State \textbf{Training:} \ $\beta \gets (\Phi^\top\Phi + \lambda I)^{-1}\Phi^\top\mathbf{y}$, \quad $\hat{\sigma}^2 \gets \tfrac{1}{N}\lVert \mathbf{y} - \Phi\beta \rVert_2^2$
\State \textbf{Generation:} \ $y_t \sim \mathcal{N}(\phi_t^\top\beta,\ \hat{\sigma}^2)$
\end{algorithmic}
\end{algorithm}

\begin{algorithm}[h!]
\caption{GRU parametric simulator (autoregressive Gaussian emission; hidden $128$).}
\label{alg:gru}
\begin{algorithmic}[1]
\Require weekly sequences $\{(\phi_t, y_t)\}_{t=1}^{168}$
\State $r_t \gets \mathrm{GRU}_\theta(r_{t-1}, \phi_t)$, \quad $(\mu_t, \sigma_t) \gets g_\theta(r_t)$
\State \textbf{Training:} \ $\theta \gets \operatorname*{arg\,min}\limits_{\theta} \sum_{\mathrm{weeks}} \sum_{t=1}^{168} \big[ \tfrac{(y_t - \mu_t)^2}{2\sigma_t^2} + \log\sigma_t \big]$, \ using data $y_{t-1}$ in $\phi_t$ \Comment{teacher forcing}
\State \textbf{Generation:} \ \textbf{for} $t=1,2,\dots$: \ $y_t \sim \mathcal{N}(\mu_t, \sigma_t^2)$, then set $\phi_{t+1}$ from the sampled $y_t$ \Comment{free-running}
\end{algorithmic}
\end{algorithm}

The GRU trains with Adam for $12$ epochs (learning rate $10^{-3}$, weight decay $10^{-5}$) on $168$-hour chunks (batch $128$, gradient clip $5$).
In short, the retrieval sampler PSRS matches the hourly marginal and short-range autocorrelation by splicing real transitions, but, having no participant-level signal once the survey is removed, it collapses between-participant heterogeneity (bu-acf and bu-run well below $1$); KNN averages over neighbors and under-disperses further.
The linear SEM recovers short-range dynamics, but, sampling a Gaussian on the log scale and feeding it back, it accumulates variance under closed-loop rollout, drifts, and over-disperses (a large daily-profile Wasserstein), which is why its raw autocorrelation error is deceptively low even as its distributional fit is among the worst.
The GRU is far stronger.
Scored on the identical footprint-matched protocol as Table~\ref{tab:eval_main}, at the second hand-off (v2/v3 $\to$ v4) it reaches a daily-profile Wasserstein of $9.28 \pm 0.13$, ACF-MAE $0.064 \pm 0.004$, run-length Wasserstein $0.80 \pm 0.06$, bu-acf $0.21 \pm 0.03$, and bu-run $0.13 \pm 0.02$; at the first hand-off (AoU $\to$ v2/v3) the corresponding values are $10.00 \pm 0.06$, $0.196 \pm 0.003$, $1.19 \pm 0.04$, $0.25 \pm 0.02$, and $0.23 \pm 0.01$.
Its recurrent state gives it the best short-range dynamics of any baseline at the second hand-off, but with no participant-level input its heterogeneity ratios stay far below one, and its daily-profile fit never approaches the fine-tuned twin's $8.43 \pm 0.19$.
None of the baselines reproduce the multi-scale temporal structure together with the between-participant heterogeneity that the survey-conditioned diffusion twin captures.

\section{Attributing credit across training stages}
\label{appendix:credit}

The twin is built in three stages: pretraining on \emph{All of Us}, fine-tuning on the prior HeartSteps deployments (v2 and v3), and inference-time calibration toward the upcoming deployment (Section~\ref{sec:llm}).
A natural question is how much each stage contributes to the fidelity reported in Table~\ref{tab:eval_main}, and in particular whether fine-tuning earns its place given that the learned proximal effect is close to null (Section~\ref{sec:action_cond}).
We isolate each stage with an ablation that crosses the training stage (pretraining only; with the v2/v3 fine-tune; with a v4 fine-tune) against calibration (absent or present), a $3\times2$ grid (Table~\ref{tab:credit}).
The distributional column is the same daily-profile Wasserstein as Table~\ref{tab:eval_main} (footprint-matched against the full v4 cohort at $N=150$); the remaining columns are computed on a held-out v4 test split with per-participant matching.
The pretrain-only row is the fully trained 150,000-iteration stage-1 checkpoint from which the fine-tuned models in the main text are adapted, so the rows trace a single lineage; Table~\ref{tab:eval_main}'s pretrain-only row instead reports the 70,000-iteration checkpoint used for direct generation (Appendix~\ref{sec:pretrain}), which is why the two daily-Wasserstein values differ ($9.93$ here vs.\ $9.60$ there).
Calibration is applied with the same scientist-and-LLM calibration target $\vmu^\star$ and one common trust level ($\gamma = 0.9$) in every calibrated row, so that within each pair the only change is the calibration step; Table~\ref{tab:eval_main}'s calibrated row uses the $\gamma = 1.5$ default of Section~\ref{sec:reward_tilt}, and the daily-profile Wasserstein is insensitive across this range (Appendix~\ref{appendix:calibration}).
The two v4-fine-tuned rows are the \emph{oracle}, the twin fine-tuned on the held-out v4 data; because it sees that data, it bounds achievable fidelity rather than naming a model available before the deployment.

\begin{table}[h]
\centering
\footnotesize
\setlength{\tabcolsep}{3pt}
\caption{Credit attribution across training stages on HS v4.
The daily W column is the daily-profile Wasserstein of Table~\ref{tab:eval_main}
(footprint-matched against the full v4 cohort, $N=150$); the remaining columns
are computed on a held-out v4 test split with per-participant matching. Each
training stage (pretrain, plus v2/v3 fine-tune (FT), plus v4 fine-tune) is
shown without and with calibration. First three columns lower-is-better; last
two are closer-to-1-better. Calibration uses the same scientist-and-LLM
target and a common trust level ($\gamma=0.9$) throughout; the pretrain rows use
the 150,000-iteration checkpoint the fine-tuned models are adapted from, not the
70,000-iteration checkpoint of Table~\ref{tab:eval_main}. Bold marks the best no-v4-data configuration on the three
lower-is-better metrics; the v4-fine-tune rows are the oracle, which
uses held-out v4 data and bounds achievable fidelity.}
\label{tab:credit}
\newcommand{\pmse}[1]{\ensuremath{{\scriptstyle\,(\pm #1)}}}
\resizebox{\textwidth}{!}{%
\begin{tabular}{l|ccccc}
\toprule
Stage & Daily W (log) $\downarrow$ &
  ACF-MAE short $\downarrow$ & Run-length W $\downarrow$ &
  bu-acf $\to 1$ & bu-run $\to 1$ \\
\midrule
AoU pretrain (no FT)           & 9.93\pmse{0.22}  & 0.087\pmse{0.018} & 1.81\pmse{0.38} & 0.64\pmse{0.19} & 0.18\pmse{0.03} \\
AoU pretrain + calibration     & 9.31\pmse{0.20} & 0.101\pmse{0.020} & 1.38\pmse{0.43} & 0.68\pmse{0.23} & 0.28\pmse{0.06} \\
\midrule
+ v23 fine-tune                & 8.43\pmse{0.19}   & 0.039\pmse{0.016} & 0.74\pmse{0.38} & 0.45\pmse{0.18} & 0.46\pmse{0.14} \\
+ v23 fine-tune + calibration  & \textbf{8.33}\pmse{0.17} & \textbf{0.031}\pmse{0.012} & \textbf{0.36}\pmse{0.28} & 0.42\pmse{0.17} & 0.51\pmse{0.12} \\
\midrule
v4 oracle (FT on v4)           & 8.44\pmse{0.18}   & 0.062\pmse{0.015} & 0.48\pmse{0.35} & 0.63\pmse{0.13} & 0.62\pmse{0.16} \\
v4 oracle + calibration        & 8.38\pmse{0.21}   & 0.044\pmse{0.017} & 0.27\pmse{0.28} & 0.54\pmse{0.21} & 0.57\pmse{0.14} \\
\bottomrule
\end{tabular}}
\end{table}

We read three things from the table.
First, fine-tuning is what supplies temporal structure and well-scaled heterogeneity.
The pretrain-only twin has seen no HeartSteps data, and although it conditions on the baseline survey it reproduces the v4 daily profile poorly (daily Wasserstein $9.93$, far above the fine-tuned twin's $8.43$), produces incoherent activity bouts (run-length Wasserstein $1.81$), and gets the between-participant spreads wrong in both directions, over-dispersing participant means while collapsing the spread of active-run lengths (bu-run $0.18$).
Adding the v2/v3 fine-tune cuts the daily Wasserstein to $8.43$, more than halves the run-length Wasserstein, and lifts the run-length heterogeneity ratio from $0.18$ toward v4 ($0.46$).
None of these gains run through the intervention channel, whose learned effect is almost null (Section~\ref{sec:action_cond}); they come from the fine-tune adapting the baseline, no-nudge behavior to the HeartSteps population.

Second, calibration improves the daily-profile fit on every base, but it cannot substitute for fine-tuning.
The largest absolute gain is on the pretrain base ($9.93 \to 9.31$), where the residual gap to v4 is largest, yet even calibrated the pretrain twin remains far from the fine-tuned rows ($9.31$ against $8.33$); on the fine-tuned bases the gain is small because little gap remains ($8.43 \to 8.33$ on the v2/v3 twin, $8.44 \to 8.38$ on the v4 oracle), with the per-hour profile sharpened in each case.
A per-hour mean shift, however aggressive, cannot supply the within-day structure that fine-tuning provides.

Third, what calibration cannot do, on any base, is manufacture between-participant heterogeneity.
The bu-acf and bu-run ratios are set by the base and move little under calibration (v4 oracle $0.63 \to 0.54$ and $0.62 \to 0.57$; v2/v3 twin $0.45 \to 0.42$ and $0.46 \to 0.51$), because a shift toward a single shared per-hour target cannot change how participants differ from one another.
That heterogeneity, together with the temporal structure, is what fine-tuning supplies: the fine-tune is what lifts the run-length heterogeneity ratio from the pretrain twin's $0.18$ toward v4 ($0.46$) and brings the activity-bout structure into range.
The two stages are therefore complementary, not interchangeable: fine-tuning produces the temporal and between-participant structure that the downstream design decisions turn on (Section~\ref{sec:discussion}), calibration sets the marginal level on top of it, and calibration without fine-tuning closes only a fraction of the gap.
The intervention channel contributes little to either, because the program's real proximal effect is weak (Section~\ref{sec:action_cond}).

\subsection*{Data Repository / Code}
The \emph{All of Us} data used for pre-training are available to registered researchers through the NIH \emph{All of Us} Researcher Workbench (\url{https://www.researchallofus.org}); per the program's data-use policies, participant-level data cannot be shared outside the Workbench.
The HeartSteps v2--v4 data are not publicly available due to participant-privacy restrictions; access may be requested from the HeartSteps investigators.
Code implementing the twin architecture, the calibration sampler, and the evaluation pipeline will be made publicly available upon publication.

\ifhdsrnatbib
  \bibliographystyle{apalike}
  \bibliography{references}
\else
  \printbibliography
\fi

\end{document}

%% file: macros.tex
\usepackage{amsmath,bbm,bm}
\usepackage{amsfonts}
\usepackage{amsthm}
\usepackage{mathtools}
\usepackage{algorithm}
\usepackage{algorithmicx}
\usepackage{algpseudocode}
\usepackage{tabularx}
\algdef{SE}[SUBALG]{Indent}{EndIndent}{}{\algorithmicend\ }%
\algtext*{Indent}
\algtext*{EndIndent}

\usetikzlibrary{positioning,arrows.meta,calc,fit,shapes.geometric,backgrounds,decorations.pathreplacing}

\usepackage{xcolor}

\definecolor{promptframe}{rgb}{0.15,0.20,0.45}
\definecolor{promptback}{rgb}{0.93,0.95,1.0}
\newsavebox{\promptboxsave}
\newenvironment{promptbox}[1][]{%
  \setlength{\fboxsep}{7pt}%
  \begin{lrbox}{\promptboxsave}%
  \begin{minipage}{\dimexpr\textwidth-2\fboxsep-2\fboxrule\relax}%
  \sloppy
  \def\promptboxtitle{#1}%
  \ifx\promptboxtitle\empty\else
    {\bfseries\small\color{promptframe}\promptboxtitle}\par\smallskip
  \fi
}{%
  \end{minipage}%
  \end{lrbox}%
  \par\noindent\fcolorbox{promptframe}{promptback}{\usebox{\promptboxsave}}\par
}
\newfloat{promptfloat}{htbp}{lopr}
\floatname{promptfloat}{Prompt}

\newcount\comments  
\newcommand{\genComment}[2]{\ifnum\comments=1{\textcolor{#1}{\textsf{\footnotesize #2}}}\fi}

\def\eqref#1{equation~\ref{#1}}

\def\1{\bm{1}}

\def\eps{{\epsilon}}

\def\vzero{{\bm{0}}}

\def\vmu{{\bm{\mu}}}

\def\vh{{\bm{h}}}

\def\vs{{\bm{s}}}

\def\vv{{\bm{v}}}

\def\vx{{\bm{x}}}

\def\mA{{\bm{A}}}
\def\mB{{\bm{B}}}

\def\mI{{\bm{I}}}

\DeclareMathAlphabet{\mathsfit}{\encodingdefault}{\sfdefault}{m}{sl}
\SetMathAlphabet{\mathsfit}{bold}{\encodingdefault}{\sfdefault}{bx}{n}

\def\gL{{\mathcal{L}}}

\def\gN{{\mathcal{N}}}

\def\sR{{\mathbb{R}}}



%% file: Figure/eval/main_table.tex
\begin{table}[h]
\centering
\footnotesize
\setlength{\tabcolsep}{2.5pt}
\caption{JITAI-Twin fidelity across the two pretend-prospective hand-offs.
\textbf{(a)} The AoU-pretrained twin and baselines evaluated on HS v2/v3; \textbf{(b)} the v2/v3-fine-tuned twin and v2/v3-trained baselines evaluated on HS v4.}
\label{tab:eval_main}
\newcommand{\pmse}[1]{\ensuremath{{\scriptstyle\,(\pm #1)}}}

\textbf{(a) Trained on \emph{All of Us}, evaluated on all of HS v2/v3}\\[3pt]
\begin{tabular}{l|ccccc}
\toprule
Method & Daily W (log) $\downarrow$ &
  ACF-MAE (log) $\downarrow$ & Run-length W $\downarrow$ &
  bu-acf $\to 1$ & bu-run $\to 1$ \\
\midrule
Diffusion (AoU pretrain)  & 10.82\pmse{0.22} & 0.272\pmse{0.016} & 2.34\pmse{0.17} & \textbf{0.44}\pmse{0.05} & 0.20\pmse{0.04} \\
\quad + \textit{calibration} & 10.22\pmse{0.26} & 0.249\pmse{0.016} & 1.94\pmse{0.18} & \textbf{0.41}\pmse{0.07} & 0.25\pmse{0.05} \\
\quad + \textit{v23 fine-tune oracle}$^{\dagger}$ & \textbf{8.43}\pmse{0.21} & 0.115\pmse{0.013} & 0.98\pmse{0.21} & 0.26\pmse{0.04} & \textbf{0.53}\pmse{0.10} \\
\hline
KNN                          & 10.77\pmse{0.06} & \textbf{0.061}\pmse{0.005} & \textbf{0.58}\pmse{0.07} & 0.05\pmse{0.01} & 0.06\pmse{0.01} \\
PSRS                    & 9.65\pmse{0.12} & 0.096\pmse{0.010} & 0.79\pmse{0.12} & 0.17\pmse{0.01} & 0.11\pmse{0.02} \\
linear SEM                   & 10.85\pmse{0.07} & 0.133\pmse{0.010} & 1.67\pmse{0.15} & 0.14\pmse{0.01} & 0.07\pmse{0.01} \\
\midrule
\emph{AoU real data}$^{\ddagger}$          & 8.45\pmse{0.07} & 0.054\pmse{0.012} & 0.39\pmse{0.15} & 0.96\pmse{0.07} & 0.73\pmse{0.11} \\
\emph{HS v2/v3 real data}$^{\S}$          & 7.93\pmse{0.10} & 0.011\pmse{0.014} & 0.19\pmse{0.17} & 1.00\pmse{0.14} & 1.06\pmse{0.10} \\
\bottomrule
\end{tabular}

\vspace{8pt}

\textbf{(b) Trained on AoU, fine-tuned on HS v2/v3, evaluated on all of HS v4}\\[3pt]
\begin{tabular}{l|ccccc}
\toprule
Method & Daily W (log) $\downarrow$ &
  ACF-MAE (log) $\downarrow$ & Run-length W $\downarrow$ &
  bu-acf $\to 1$ & bu-run $\to 1$ \\
\midrule
Diffusion (AoU pretrain)  & 9.60\pmse{0.22} & 0.122\pmse{0.023} & 1.89\pmse{0.21} & \textbf{0.43}\pmse{0.07} & \textbf{0.13}\pmse{0.07} \\
Diffusion (v23 fine-tune) & \textbf{8.43}\pmse{0.19} & \textbf{0.045}\pmse{0.012} & 0.80\pmse{0.24} & 0.30\pmse{0.05} & \textbf{0.25}\pmse{0.14} \\
\quad + \textit{calibration} & \textbf{8.34}\pmse{0.18} & \textbf{0.071}\pmse{0.015} & \textbf{0.22}\pmse{0.13} & 0.28\pmse{0.05} & \textbf{0.31}\pmse{0.15} \\
 \quad + \textit{v4 fine-tune oracle}$^{\dagger}$  & \textbf{8.44}\pmse{0.18} & \textbf{0.049}\pmse{0.011} & \textbf{0.58}\pmse{0.24} & \textbf{0.38}\pmse{0.07} & \textbf{0.33}\pmse{0.18} \\
\hline
KNN                          & \textbf{8.62}\pmse{0.18} & 0.223\pmse{0.012} & 0.52\pmse{0.07}  & 0.04\pmse{0.01} & 0.07\pmse{0.03} \\
PSRS                    & 10.81\pmse{0.34} & 0.145\pmse{0.017} & 0.70\pmse{0.13} & 0.10\pmse{0.02} & 0.07\pmse{0.04} \\
linear SEM                   & 10.17\pmse{0.19} & 0.091\pmse{0.014} & 0.91\pmse{0.19} & 0.12\pmse{0.02} & 0.03\pmse{0.02} \\
\midrule
\emph{HS v2/v3 real data}$^{\ddagger}$     & 9.24\pmse{0.16} & 0.154\pmse{0.021} & 0.66\pmse{0.22} & 0.95\pmse{0.11} & 0.64\pmse{0.31} \\
\emph{HS v4 real data}$^{\S}$             & 8.30\pmse{0.18} & 0.023\pmse{0.020} & 0.05\pmse{0.05} & 1.06\pmse{0.24} & 1.04\pmse{0.13} \\
\bottomrule
\end{tabular}

\vspace{6pt}
\begin{minipage}{\linewidth}
\footnotesize\raggedright
\emph{Note:} Columns 1--3 are lower-is-better, the last two closer-to-1-better; $\pm x$ is one bootstrap standard error. Bold marks every non-reference entry whose $\pm$ 1 SE overlaps with that of the column's best non-reference entry. Oracle rows ($^{\dagger}$) are fine-tuned on the target data. The last two reference rows are the metrics computed on the real prior-deployment data ($^{\ddagger}$) and the real target data ($^{\S}$).
\end{minipage}
\end{table}

%% file: appendix_model.tex
\section{Diffusion-Based JITAI-Twin Architecture}
\label{sec:architecture_detail}

This appendix specifies the twin's diffusion architecture in full: the backbone, its temporally consistent (forward-only) design, survey conditioning, the log-compressed step representation, and hyperparameters.

The JITAI-Twin denoiser is built on the TimeWeaver-SSSD backbone
\citep{narasimhan2024time}, which adapts the SSSD-S4 state-space
diffusion architecture \citep{alcaraz2022diffusion} to heterogeneous
time-series conditioning.
In both prior works the conditioning signal is the observed portion of
the target series itself: given a partially masked time series, the network
imputes the missing values by conditioning on the non-masked channels.
Our digital-twin setting differs in two fundamental respects, motivating
the architectural changes described in this section.

\paragraph{Survey conditioning}
Our model moves away from masking-based imputation to a cross-attention-based
conditioning architecture, as the model should condition on heterogeneous survey
data obtained at participant enrollment.
We replace the masking path with per-block cross-attention over a frozen T5
token sequence derived from a natural-language summary of the baseline survey
\citep{raffel2020t5}.
This lets the model attend to specific survey items at each residual block
and adapts to new survey schemas without structural changes, as the
harmonization operates at the text level (Appendix~\ref{appendix:embedding}).

\paragraph{Intervention conditioning}
The intervention sequence is a separate, time-ordered binary signal
distinct from the step-count trajectory.
In the original TimeWeaver, conditioning channels are processed symmetrically
in time. We instead introduce a dedicated temporally consistent intervention embedding module
that is entirely absent at pre-training and is zero-initialized at the start
of fine-tuning (Section~\ref{sec:action_embed}).

All other aspects of the TimeWeaver-SSSD backbone, namely the WaveNet-style
residual stack, the structured state-space (S4) temporal-mixing layers, the
diffusion noise schedule, and the skip-summed output projection, are
retained unchanged.

\subsection{Diffusion Framework}
\label{sec:diffusion}

 We adopt the denoising diffusion probabilistic model (DDPM) framework
\citep{ho2020denoising} with $\varepsilon$-prediction.
In the forward process, clean data is progressively corrupted by Gaussian noise:
\[
  q(\vx_t \mid \vx_{t-1})
  = \gN\!\bigl(\vx_t;\,
      \sqrt{1-\beta_t}\,\vx_{t-1},\;
      \beta_t \mI\bigr),
  \qquad t = 1, \ldots, T.
\]
Here, we define $F = 1$ and $L = 168$ representing a week-long window of log-normalized hourly step count.
The learned reverse process recovers $\vx_0$ from pure noise $\vx_T \sim \gN(\vzero, \mI)$:
\[
  p_\theta(\vx_{t-1} \mid \vx_t, C_i, A)
  = \gN\!\bigl(\vx_{t-1};\,
      \vmu_\theta(\vx_t, t, C_i, A),\;
      \sigma_t^2 \mI\bigr),
\]
where $C_i$ is the participant's T5 token embedding and $A \in \{0,1\}^L$
is the week's intervention sequence.
Training minimizes the simplified noise-prediction objective:
\begin{equation}
  \gL(\theta)
  = \E_{t,\,\vx_0,\,\eps}
    \bigl\|\eps - \eps_\theta(\vx_t, t, C_i, A)\bigr\|_{\mathcal{M}}^2,
  \label{eq:ddpm_loss}
\end{equation}
where $\mathcal{M}$ is a binary mask that restricts the MSE to hours with
valid Fitbit wear data.  
The key departure from typical DDPM applications is that the model generates
a complete week jointly rather than autoregressively one hour at a time.
Temporal consistency is enforced architecturally, not by conditioning
each step on all previous ones (see Section~\ref{sec:temporal_consistency}).

\subsection{Network Stack}
\label{sec:stack}

\begin{figure}[h!]
\centering
\includegraphics[width=\linewidth]{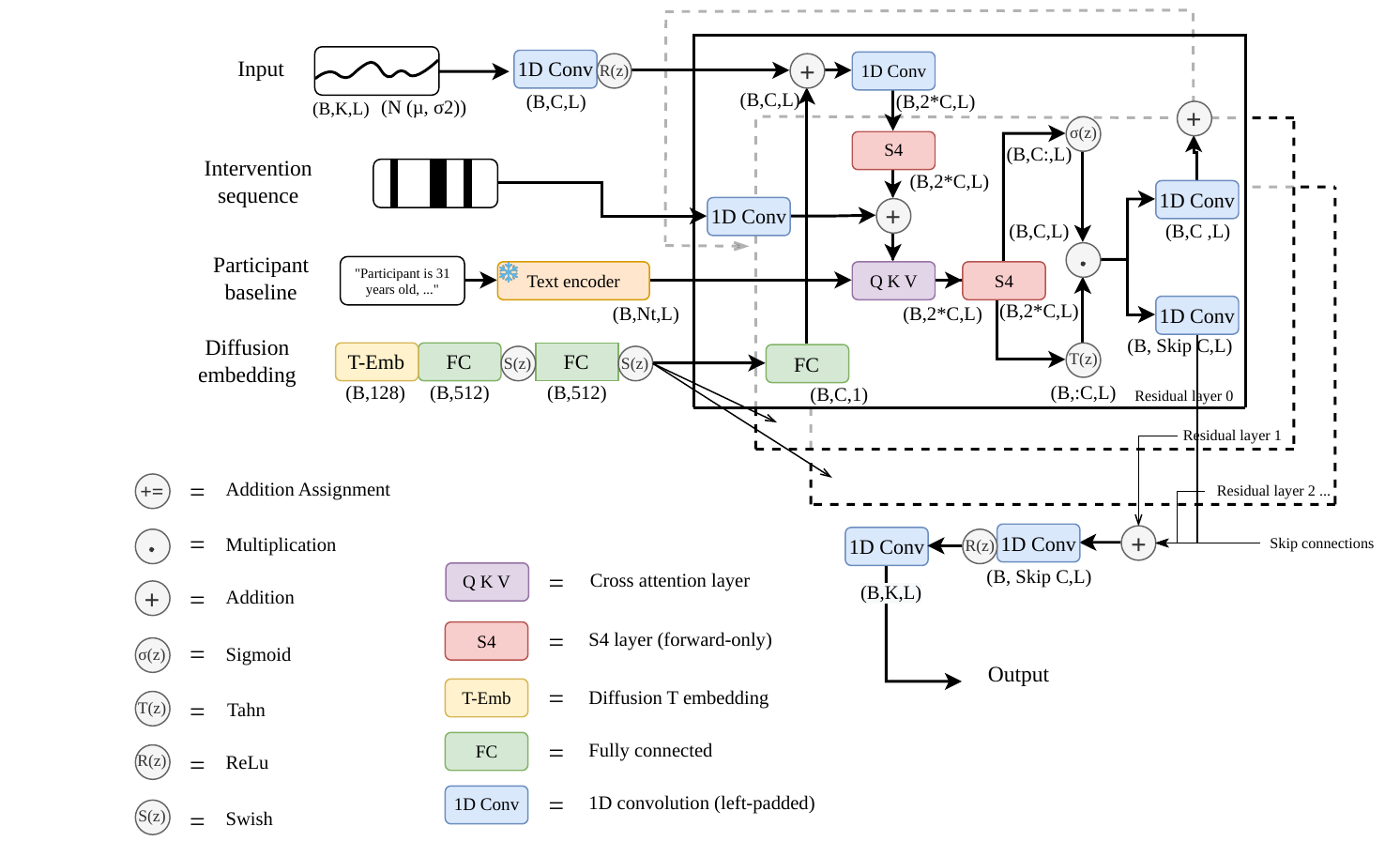}
\caption{Architecture of the JITAI-Twin denoising network $\eps_\theta$.
The noisy week-long window enters a stack of 36 identical residual blocks
(one shown in detail, with two more drawn faded behind it).
Each block applies two forward-only S4 temporal mixers, cross-attention that
conditions on the participant's frozen T5 baseline embedding $C_i$ (red
dashed box), and a tanh--sigmoid gate, emitting a hidden state for the next
block (top-right $+$) and a skip output.
The 36 skip outputs are summed (bottom $+$) and projected back to one channel
by two $1\times1$ convolutions.
Temporal consistency is enforced by two mechanisms (Section~\ref{sec:temporal_consistency}):
left-padded temporally consistent convolutions and forward-only S4 kernels, so the hour-$t$
prediction never depends on a future hour or a future action.
At fine-tuning a zero-initialized temporally consistent intervention-embedding module
(Section~\ref{sec:action_embed}) injects the action sequence additively after
the first S4 layer.}
\label{fig:denoising_block}
\end{figure}

The denoiser $\eps_\theta$, summarized at the block level in
Figure~\ref{fig:denoising_block}, consists of three components.
An \textbf{input projection}, a point-wise 1-D convolution with a
rectifier nonlinearity, maps the single-channel noisy step window into a
256-channel hidden representation.
A stack of \textbf{36 residual blocks} (Section~\ref{sec:resblock})
transforms this representation in place while emitting a 256-channel skip
contribution at each block. The skip contributions are summed across all 36
blocks and scaled by $1/\!\sqrt{36}$.
An \textbf{output decoder}, two stacked point-wise 1-D convolutions
(the second zero-initialized), projects the summed skip back to a
single-channel noise prediction $\hat\eps \in \sR^{1 \times 168}$.
Zero-initializing the final convolution ensures the network predicts exactly
zero noise at the start of training, which empirically stabilizes DDPM
convergence.
The pre-training model has 83.2M parameters, all trained during pre-training.
Fine-tuning updates only a ${\approx}1.6\%$ subset (Section~\ref{sec:peft}).

\subsection{Residual Block}
\label{sec:resblock}

Each of the 36 residual blocks $f^{(\ell)}_\theta: \vh \mapsto \vh'$ applies
the following operations sequentially to its 256-channel input
$\vh \in \sR^{B \times 256 \times 168}$.
Step~4 below is present only at fine-tuning. During pre-training the
intervention module does not exist and the block proceeds directly from
step~3 to step~5.

\begin{enumerate}
  \item \textbf{Diffusion-step injection.}
    A per-block linear projection maps the 512-dimensional diffusion-step
    embedding to 256 dimensions and adds the result to $\vh$ via
    broadcasting over the time axis.

  \item \textbf{Temporally consistent 1-D convolution.}
    A 1-D convolution (256 input channels, 512 output channels, kernel
    size~3) with left-only padding expands the channel dimension while
    enforcing that the output at hour $t$ depends only on inputs at hours
    $\leq t$ (Section~\ref{sec:temporal_consistency}).

  \item \textbf{First temporally consistent state-space layer.}
    A forward-only S4 layer (Section~\ref{sec:temporal_consistency}) with hidden width
    512, HiPPO state dimension 64, and sequence length $L_{\max} = 168$
    mixes information across time in a temporally consistent, parameter-efficient manner
    via an $O(L \log L)$ FFT-based convolution.

  \item \textbf{Intervention injection~(\textit{fine-tuning only}).}
    The temporally consistent intervention embedding module (Section~\ref{sec:action_embed})
    produces a 512-channel additive shift from the binary intervention
    sequence $A$, which is added to the hidden state:
    $\vh \leftarrow \vh + \operatorname{Embed}_{\mathrm{tc}}(A)$.

  \item \textbf{Survey cross-attention.}
    The hidden state $\vh$ serves as the query matrix, and the participant's
    T5 token sequence $C_i$ provides the keys and values.
    Scaled-dot-product attention with 8 heads and 64 dimensions per head
    injects per-participant identity at each block.
    An attention mask prevents attention to padding tokens.
    Because attention operates over the survey-token axis rather than
    the time axis, this step does not carry information across time steps and is temporally consistent (Section~\ref{sec:temporal_consistency}).

  \item \textbf{Second temporally consistent state-space layer.}
    A second S4 layer with the same hyperparameters as step~3 refines the
    representation after the survey conditioning.

  \item \textbf{Gated activation.}
    The 512-channel hidden state is split equally along the channel axis and
    combined as $\tanh(\vh_{:256}) \odot \sigma(\vh_{256:})$, a WaveNet-style
    multiplicative gate that introduces nonlinearity while controlling
    information flow.

  \item \textbf{Residual and skip connections.}
    A point-wise convolution produces the 256-channel residual update (added
    to the block input to form $\vh'$) and a 256-channel skip contribution
    forwarded to the output accumulator.
\end{enumerate}

\subsection{Temporal Consistency}
\label{sec:temporal_consistency}

The temporally-consistent-simulator requirement ($S_{t+1} \perp\!\!\!\perp A_{t+2},\allowbreak
A_{t+3},\allowbreak\ldots \mid \mathcal{H}_t, A_t$) is enforced architecturally
by two independent mechanisms, both of which must be active simultaneously
to eliminate temporal leakage.

\textbf{Temporally consistent 1-D convolutions.}
The 1-D convolution in step~2 of each residual block uses
left-only padding: $\text{dilation} \times (\text{kernel size} - 1)$
zeros are prepended to the input, with no padding on the right.
The output at position $t$ therefore depends only on input positions
$\leq t$, whereas the symmetric padding used in the bidirectional variant
would allow each position to see $(\text{kernel size} - 1)/2$ positions
ahead.

\textbf{Forward-only state-space S4 kernels.}
Each S4 layer convolves the input with a structured kernel derived from the
HiPPO state matrix, computed efficiently via FFT.
The bidirectional variant of this convolution concatenates a forward kernel
with a time-reversed backward kernel, giving every output position access
to the full sequence.
With the backward kernel disabled, only the forward direction is active:
the output at position $t$ depends only on positions $\leq t$.

Together these two mechanisms make every data-to-data path in the denoiser
left-to-right in time.
The diffusion-step injection (step~1) and the gated activation (step~7) are
pointwise in time and contribute no leakage.
The survey cross-attention (step~5) operates over survey tokens, not time
steps, and is therefore also exempt.

\subsection{Survey Conditioning via Cross-Attention}
\label{sec:survey_cond}

Each participant's harmonized survey responses are rewritten as a short
natural-language summary and encoded into a token sequence
$C_i \in \sR^{B \times M_i \times d_c}$ with $d_c = 512$ and a variable
number of tokens $M_i$ per participant, padded to a common maximum length
within each mini-batch. 
The text embedding is obtained using the T5 text transformer \citep{raffel2020t5}, which remains frozen throughout all training
stages.

At every residual block the 512-dimensional hidden state serves as the
query matrix ($[B, L, 512]$) and the T5 token sequence provides the keys
and values ($[B, M_i, 512]$).
Multi-head attention ($8$ heads, $64$ dimensions per head) with a binary
attention mask on padding positions updates $\vh$ via a residual connection.
Using the full token sequence rather than a single pooled vector preserves
the item-level structure of the survey and allows the model to attend to
different survey fields at different blocks, which we find empirically
beneficial for hour-of-day-specific generation, outperforming a pooled-embedding
FiLM variant on temporal-structure metrics in our full evaluation sweep
(reported in the supplementary material).

Two properties of this design are worth noting.
First, because cross-attention is computed over the survey-token axis and
not over the time axis, no information flows across hours via this path.
Temporal consistency is unaffected.
Second, because the T5 encoder is frozen, its weights do not change during
either pre-training or fine-tuning. Only the cross-attention projection
matrices in the residual blocks adapt.

\subsection{Log-Compressed Step Representation}
\label{sec:linlog}

Hourly step counts have a heavy-tailed, zero-inflated marginal distribution
that is poorly suited to direct Gaussian diffusion.
We apply a piecewise log-compression before training: writing $a = 600$ for the breakpoint, values $x \le a$ map to $x/a$ and values $x > a$ map to $1 + \log_{10}(x/a)$, a continuous transform that is linear below the breakpoint and logarithmic above it.
The resulting representation lies in a compact range ($[0, \approx 2]$ for typical hourly counts)
that stabilizes training and reflects the perceptual scale of human
activity.
All model inputs, outputs, and diffusion noise are in this compressed space.
Evaluation metrics are reported after inverting the transform back to raw
step counts. The inverse is exponential above the threshold, so small
deviations in the compressed domain correspond to large differences at the
tail of the step distribution.
All calibration computations (Appendix~\ref{appendix:calibration}) are likewise
carried out in the compressed space, with a per-hour scaling factor applied
to convert the practitioner's raw-step target into the model's native
coordinates (Appendix~\ref{appendix:llm_calib}).

\subsection{Architecture Hyperparameters}
\label{sec:arch_hparams}

Table~\ref{tab:hyperparams} lists the architecture hyperparameters of the
1-week ($L = 168$) temporally consistent denoising network. The optimization and
training-stage settings are in Table~\ref{tab:hparam_summary}.

\begin{table}[h!]
\centering
\small
\caption{Architecture hyperparameters of the JITAI-Twin denoising network.}
\label{tab:hyperparams}
\begin{tabular}{l|l}
\hline
Window length, $L$ & 168 \\
Input/output channels, $F$ & $1$ \\
Hidden (residual) width & 256 \\
Skip-connection width & 256 \\
Number of residual blocks & 36 \\
S4 layer state dimension & 64 \\
S4 layer max sequence length & 168 \\
S4 dropout / layer-norm & 0.0 / on \\
S4 layer bidirectional & False \\
Cross-attention heads $\times$ dim/head & $8 \times 64$ \\
Diffusion-step embedding (in / mid / out) & 128 / 512 / 512 \\
Diffusion steps $T$ & 200 \\
Noise schedule & linear $\beta_0 = 10^{-4} \to \beta_T = 0.02$ \\
\hline
\end{tabular}
\end{table}

\subsection{Ablations: Bidirectional and FiLM-Conditioned Variants}
\label{sec:ablations}

Two ablation configurations each differ from the model above by a single choice.

\textbf{Bidirectional variant.}
Setting the S4 layers to bidirectional adds the time-reversed S4 kernel
(Section~\ref{sec:temporal_consistency}), so the receptive field spans the whole window
and the model is no longer a valid temporally consistent generative model for use as a
JITAI-Twin. We retain it only as an ``oracle'' upper bound on fitting quality.

\textbf{FiLM-conditioned variant.}
Replaces the per-block cross-attention with a per-block FiLM layer
$\vh \leftarrow (\mathbf{1} + \bm{\gamma}) \odot \vh + \bm{\beta}$, where
$(\bm{\gamma}, \bm{\beta})$ come from a linear projection of the pooled T5
embedding. This is more parameter-efficient than cross-attention but discards
the token-level structure of the T5 embedding. In the same evaluation sweep
it underperforms the cross-attention variant on temporal-structure metrics,
motivating the attention-based design used in the main paper.

%% file: appendix_train_peft.tex
\section{Stage 1: Pre-training on \emph{All of Us}}
\label{sec:pretrain}

Pre-training uses the NIH \emph{All of Us} Research Program \citep{ramirez2022all}, a national observational cohort in which participants wear activity trackers in daily life without receiving any intervention. 
The participant survey consists of two parts administered at the early stage of All of Us enrollment \citep{ramirez2022all}. The part 1 survey collects participant ``Basics'', ``Overall health'', and ``Lifestyle''. Part 2 focuses on healthcare, including ``Healthcare Access \& Utilization'', ``Family History'', and ``Personal Medical History''. The part 2 survey is distributed 90 days after the enrollment. The study may also distribute additional optional surveys (e.g., a survey during the COVID-19 pandemic).
We filter All of Us participants based on the following criteria: 1) who have four or more weeks of Fitbit data, and 2) who have completed all $51$ filtered survey questions used by the participant embedding (Appendix~\ref{appendix:embedding}).
This results in a cohort of $3{,}097$ eligible participants wearing Fitbits and answering all related survey questions. 
For the purposes of this study, we draw survey answers exclusively from the initial screening survey due to the high missing rates observed in follow-up surveys (e.g., physical-activity items have missing rates as high as ${\sim}73\%$).
We derive participant age from the electronic-health-record date of birth and the survey date.
To handle missingness within the enrollment period we remove implausible step values (hourly counts above $45{,}000$ steps)
as device artifacts and derive a per-hour data-validity mask from daily wear (days whose total step count does not exceed $200$ are treated as non-wear).
The validity mask restricts the diffusion loss, so the model learns actual activity dynamics rather than recording artifacts.

Because All of Us participants are not randomized to actions (i.e.,  contextually tailored activity suggestions), the goal of pre-training is to learn baseline activity dynamics: how people move
through their day, conditional on who they are, without any signal about how their behavior responds to actions.
Each participant's full step count history is tiled into 168-hour (one-week)
windows via a sliding stride.
The reported twin is pre-trained on the pre-pandemic window of this cohort
($1{,}507$ participants, $34{,}912$ week-long windows).
Steps are log-compressed (Section~\ref{sec:linlog}) before training.
The denoising loss is masked to hours with valid sensor wear:
\[
  \gL_{\mathrm{pre}}(\theta)
  = \E_{t,\,\vx_0,\,\eps}
    \bigl\|\eps - \eps_\theta(\vx_t, t, C_i)\bigr\|_{\mathcal{M}}^2,
\]
where $\mathcal{M}$ is the binary wear mask and $\eps_\theta$ at this stage
takes only the survey embedding $C_i$ as conditioning (no intervention
sequence $A$).

Pre-training uses the Adam optimizer with learning rate $2 \times 10^{-4}$ and batch size 60, and runs for 150,000 iterations, with checkpoints saved every 1,000 iterations.
Two checkpoints of this single pre-training run are used downstream: the pretrain-only twin reported in the evaluation (Table~\ref{tab:eval_main}) is the 70,000-iteration checkpoint, which generates better for the held-out HeartSteps cohorts (the fully trained checkpoint fits the \emph{All of Us} cohort more closely but transfers less well), while fine-tuning initializes from the final 150,000-iteration checkpoint (Appendix~\ref{sec:finetune}).
The pre-trained model reproduces diurnal patterns, hour-to-hour autocorrelation, and between-participant heterogeneity as captured in the All of Us cohort.
Two limitations constrain its direct applicability to the HeartSteps task.
First, it has no representation of intervention effects due to the HeartSteps contextually tailored activity suggestions: the pre-training
corpus does not contain these suggestions, so the model has nothing
from which to learn how these activity suggestions shift subsequent activity.
Second, the All of Us cohort is observational and nationally representative,
whereas the HeartSteps trials enrolled cardiac patients (v2/v3 in Seattle, WA) and
overweight sedentary adults (v4 in Southern California). The resulting marginal step-count
distributions differ, and the uncalibrated pre-trained twin under-walks
HeartSteps participants substantially on average.
Both limitations are addressed by the fine-tuning stage.

\section{Stage 2: Fine-tuning on HeartSteps Trials}
\label{sec:finetune}

Fine-tuning is required for two distinct reasons.
First, there is no representation of intervention effects due to contextually tailored activity suggestions in the All of Us data as it is purely observational. 
Second, the step-count distributions in the HeartSteps trials differ from
 All of Us in marginal level, diurnal shape, and
between-participant variance (Appendix~\ref{appendix:eda}). The fine-tuned
model must shift the pre-trained prior to match the trial population.
In addition, because the number of HeartSteps participants is only a fraction of the All of Us cohort, we develop a parameter-efficient fine-tuning (PEFT) strategy to align the JITAI-Twin to the target population under data scarcity.

\textbf{Intervention conditioning at fine-tuning only.}
\label{sec:action_timing}
The intervention-embedding module is introduced only at fine-tuning, to model how a delivered activity suggestion shifts subsequent activity.
We omit it from pre-training entirely so that pre-training focuses on learning an uncorrupted behavioral prior.
Introducing the action channel only at fine-tuning also keeps the pre-trained backbone reusable across deployments  that would deliver different intervention nudges.
The fine-tuning data consist of the step-count trajectory as the generation target, and an intervention sequence that matches the size of the step-count trajectory that is used to condition the generation.


\textbf{Temporally consistent intervention embedding.}
\label{sec:action_embed}
A temporally consistent intervention embedding module is added to each of the 36 residual
blocks at the start of fine-tuning.
Each module is a single 1-D convolution (1 input channel, 512 output
channels, kernel size~3) with left-only padding (two zeros prepended), so
that the embedding of a suggestion delivered at hour $t$ can affect the
generated step count only at hours $t' \geq t$, matching the temporally consistent
structure of the residual block's other time-mixing operations
(Section~\ref{sec:temporal_consistency}).
The weights and bias of each module are zero-initialized, so at step~0 of
fine-tuning the intervention module contributes exactly zero to the hidden state. The
action effect grows entirely through gradient updates on the HeartSteps data.

The module is injected additively into the hidden state after the first
state-space layer in each block (step~4 in Section~\ref{sec:resblock}).
Because the injection occurs before the survey cross-attention and the
second state-space layer, the generated trajectory is not a globally
additive function of the intervention at the output level, and the calibration and evaluation pipeline is
designed to accommodate this non-additivity.

\textbf{Parameter-efficient fine-tuning.}
\label{sec:peft}
Four parameter surfaces are selectively unfrozen while all other backbone parameters remain frozen.
The intervention modules are new components absent from the pre-trained checkpoint and are trained in full.
The cross-attention projections are the only path through which participant identity reaches the generator.
Between pre-training and fine-tuning, the participant population shifts from the national All of Us cohort to the target population of the HeartSteps trials (e.g., cardiac patients for HeartSteps v2/v3),
so the survey-to-signal mapping encoded in the cross-attention weights must adapt.
We found that end-to-end fine-tuning of the cross-attention layers led to unstable training given the small HeartSteps cohorts.
We therefore apply low-rank adaptation \citep[LoRA;][]{hu2022lora} to the cross-attention layers.
For each of the query, key, value, and output projections in every cross-attention module, the weight matrix $W$ is held frozen and an additive low-rank correction $\Delta W = \mB\mA$ is trained, where $\mA \in \sR^{r \times d}$, $\mB \in \sR^{d \times r}$, and $r = 8$.
The effective update is scaled by $\alpha / r$ with $\alpha = 16$.
The matrix $\mB$ is zero-initialized, so the LoRA correction starts at zero,
the same warm-start guarantee as the intervention embedding.

The input projection and output decoder are also unfrozen.
These components govern the marginal step-count scale at the entry and exit of the network. Unfreezing them allows the fine-tuned model to match the marginal step-count distribution of the HeartSteps trial population without modifying the S4 temporal-mixing layers.

The resulting trainable parameter count is approximately 1.32M, or 1.6\% of
the backbone. 


Fine-tuning initializes from the 150,000-iteration pre-trained checkpoint.
The intervention embedding weights are zero-initialized and absent from the
checkpoint. The LoRA correction matrices are freshly initialized ($\mA$ by Kaiming uniform, $\mB$ by zeros), and all other parameters are loaded directly from the checkpoint.
Training uses the Adam optimizer with learning rate $10^{-4}$ for 20,000
iterations with the same diffusion schedule as pre-training ($T = 200$, linear noise schedule), saving a checkpoint every 1,000 iterations.
Fine-tuning on cohorts this small converges early, so the reported twins use early checkpoints: 3,000 iterations for the v2/v3-fine-tuned twin of the v4 hand-off and for the v4 oracle, and 15,000 iterations for the v2/v3 oracle of the first hand-off, the latter selected for its between-participant heterogeneity on v2/v3 data.
The training objective extends the noise-prediction loss in Section~\ref{sec:diffusion} to include the intervention sequence:
\[
  \gL_{\mathrm{ft}}(\theta)
  = \E_{t,\,\vx_0,\,\eps}
    \bigl\|\eps - \eps_\theta(\vx_t, t, C_i, A)\bigr\|_{\mathcal{M}}^2.
\]
The training corpus is the combined HeartSteps v2 and v3 data, tiled into the same 168-hour sliding windows used at pre-training. The loss is masked
to hours with valid Fitbit wear.

\subsection{Hyperparameter Summary}
\label{sec:hparam_summary}

Table~\ref{tab:hparam_summary} provides a side-by-side comparison of the
pre-training and fine-tuning configurations.

\begin{table}[h!]
\centering
\small
\caption{Hyperparameters for pre-training (All of Us) and
  fine-tuning (HeartSteps v2+v3).
  Architecture hyperparameters not listed here (hidden width, S4 state
  dimension, number of residual blocks) are identical across both stages
  and are given in Table~\ref{tab:hyperparams}.}
\label{tab:hparam_summary}
\begin{tabularx}{\linewidth}{@{}l X X@{}}
\toprule
\textbf{Setting} & \textbf{Pre-training} & \textbf{Fine-tuning} \\
\midrule
\multicolumn{3}{l}{\textit{Architecture}} \\
\quad Conditioning
  & Survey cross-attention
  & Survey cross-attention + intervention embedding \\
\quad S4 time mixing    & Forward-only              & Forward-only (frozen) \\
\quad 1-D convolutions  & Temporally consistent (left-padded) & Temporally consistent (frozen) \\
\quad Trainable parameters & 83.2M                 & $\approx$1.32M \\
\midrule
\multicolumn{3}{l}{\textit{Data}} \\
\quad Training corpus
  & All of Us, pre-pandemic ($1{,}507$ participants)
  & HeartSteps v2+v3 ($113$ participants) \\
\quad Window length $L$ & 168 hours                 & 168 hours \\
\quad Loss mask         & Fitbit wear hours         & Fitbit wear hours \\
\midrule
\multicolumn{3}{l}{\textit{Optimization}} \\
\quad Optimizer         & Adam                      & Adam \\
\quad Learning rate     & $2 \times 10^{-4}$        & $1 \times 10^{-4}$ \\
\quad Batch size        & 60                        & 20 \\
\quad Iterations        & 150,000                   & 20,000 \\
\midrule
\multicolumn{3}{l}{\textit{Low-rank adaptation (cross-attention only)}} \\
\quad Rank $r$          & ---                       & 8 \\
\quad Effective scale $\alpha/r$ & ---             & 2.0 \\
\midrule
\multicolumn{3}{l}{\textit{Diffusion schedule}} \\
\quad Reverse steps $T$ & 200                       & 200 (frozen) \\
\quad Noise schedule
  & Linear $\beta_0 = 10^{-4} \to \beta_T = 0.02$
  & Same \\
\bottomrule
\end{tabularx}
\end{table}

%% file: Figure/eval/run_length_threshold_table.tex
\begin{table}[h!]
\centering\footnotesize
\caption{Active-period run-length Wasserstein at two activity thresholds (active hour $>100$ vs.\ $>250$ steps/hour), for the Table~\ref{tab:eval_main} candidates at both hand-offs, computed under the same protocol as Table~\ref{tab:eval_main}'s run-length column. The $>100$ column repeats the run-length W of Table~\ref{tab:eval_main}; the $>250$ column is a stricter-activity robustness check. Lower is better; oracle rows ($^{\dagger}$) are unavailable before the deployment runs.}
\label{tab:run_length_threshold}
\newcommand{\pmse}[1]{\ensuremath{{\scriptstyle\,(\pm #1)}}}
\textbf{(a) AoU $\to$ held-out v2/v3}\\[3pt]
\begin{tabular}{l|cc}
\toprule
Method & Run-len.\ W ($>100$) $\downarrow$ & Run-len.\ W ($>250$) $\downarrow$ \\
\midrule
Diffusion (AoU pretrain) & 2.34\pmse{0.17} & 1.53\pmse{0.12} \\
\quad + calibration (scientist+LLM) & 1.94\pmse{0.18} & 1.27\pmse{0.14} \\
Diffusion (v23 fine-tune)$^{\dagger}$ & 0.98\pmse{0.21} & 0.86\pmse{0.15} \\
KNN & 0.58\pmse{0.07} & 1.30\pmse{0.10} \\
PSRS & 0.79\pmse{0.12} & 0.67\pmse{0.10} \\
linear SEM & 1.67\pmse{0.15} & 0.83\pmse{0.10} \\
\bottomrule
\end{tabular}
\vspace{6pt}

\textbf{(b) v2/v3 $\to$ held-out v4}\\[3pt]
\begin{tabular}{l|cc}
\toprule
Method & Run-len.\ W ($>100$) $\downarrow$ & Run-len.\ W ($>250$) $\downarrow$ \\
\midrule
Diffusion (AoU pretrain) & 1.89\pmse{0.21} & 1.11\pmse{0.12} \\
Diffusion (v23 fine-tune) & 0.80\pmse{0.24} & 0.63\pmse{0.14} \\
\quad + calibration (scientist+LLM) & 0.22\pmse{0.13} & 0.20\pmse{0.12} \\
Diffusion (v4 fine-tune)$^{\dagger}$ & 0.58\pmse{0.24} & 0.67\pmse{0.13} \\
KNN & 0.52\pmse{0.07} & 0.69\pmse{0.10} \\
PSRS & 0.70\pmse{0.13} & 0.36\pmse{0.07} \\
linear SEM & 0.91\pmse{0.19} & 0.15\pmse{0.08} \\
\bottomrule
\end{tabular}
\vspace{6pt}

\end{table}